\documentclass[10pt,twocolumn,letterpaper]{article}
\usepackage[pagenumbers]{wacv} 

\usepackage{graphicx}
\usepackage{amsmath}
\usepackage{amssymb}
\usepackage{booktabs}

\usepackage{comment}
\usepackage{multirow}
\usepackage{subcaption}
\usepackage{bbding}
\usepackage{float}
\usepackage{enumitem}
\setlist{nosep}
\usepackage[normalem]{ulem}
\usepackage{pifont}
\usepackage{ctable}
\setupctable{captionskip = -0.5ex} 
\setupctable{botcap} 
\usepackage{enumitem} 
\usepackage{caption}
\usepackage{subcaption}

\usepackage{array}
\newcolumntype{L}[1]{>{\raggedright\let\newline\\\arraybackslash\hspace{0pt}}m{#1}}
\newcolumntype{R}[1]{>{\raggedleft\let\newline\\\arraybackslash\hspace{0pt}}m{#1}}

\usepackage[pagebackref,breaklinks,colorlinks]{hyperref}

\usepackage[capitalize]{cleveref}
\crefname{section}{Sec.}{Secs.}
\Crefname{section}{Section}{Sections}
\Crefname{table}{Table}{Tables}
\crefname{table}{Tab.}{Tabs.}

\begin{document}

\title{dacl10k: Benchmark for Semantic Bridge Damage Segmentation} 
\author{Johannes Flotzinger \ \ \ \ Philipp J. Rösch \ \ \ \ Thomas Braml \\
University of the Bundeswehr Munich\\
{\tt \small \{johannes.flotzinger, philipp.roesch, thomas.braml\}@unibw.de}
}

\maketitle
\thispagestyle{empty}

\begin{abstract}
Reliably identifying reinforced concrete defects (RCDs) plays a crucial role in assessing the structural integrity, traffic safety, and long-term durability of concrete bridges, which represent the most common bridge type worldwide. Nevertheless, available datasets for the recognition of RCDs are small in terms of size and class variety, which questions their usability in real-world scenarios and their role as a benchmark.      
Our contribution to this problem is ``dacl10k'', an exceptionally diverse RCD dataset for multi-label semantic segmentation comprising 9,920 images deriving from real-world bridge inspections. dacl10k distinguishes 12 damage classes as well as 6 bridge components that play a key role in the building assessment and recommending actions, such as restoration works, traffic load limitations or bridge closures. 
In addition, we examine baseline models for dacl10k which are subsequently evaluated.
The best model achieves a mean intersection-over-union of 0.42 on the test set. 
dacl10k, along with our baselines, will be openly accessible to researchers and practitioners, representing the currently biggest dataset regarding number of images and class diversity for semantic segmentation in the bridge inspection domain.  
\end{abstract}

\section{Introduction}
\label{sec:introduction}

\begin{figure*}
    \begin{center}
    \includegraphics[width=1\textwidth]{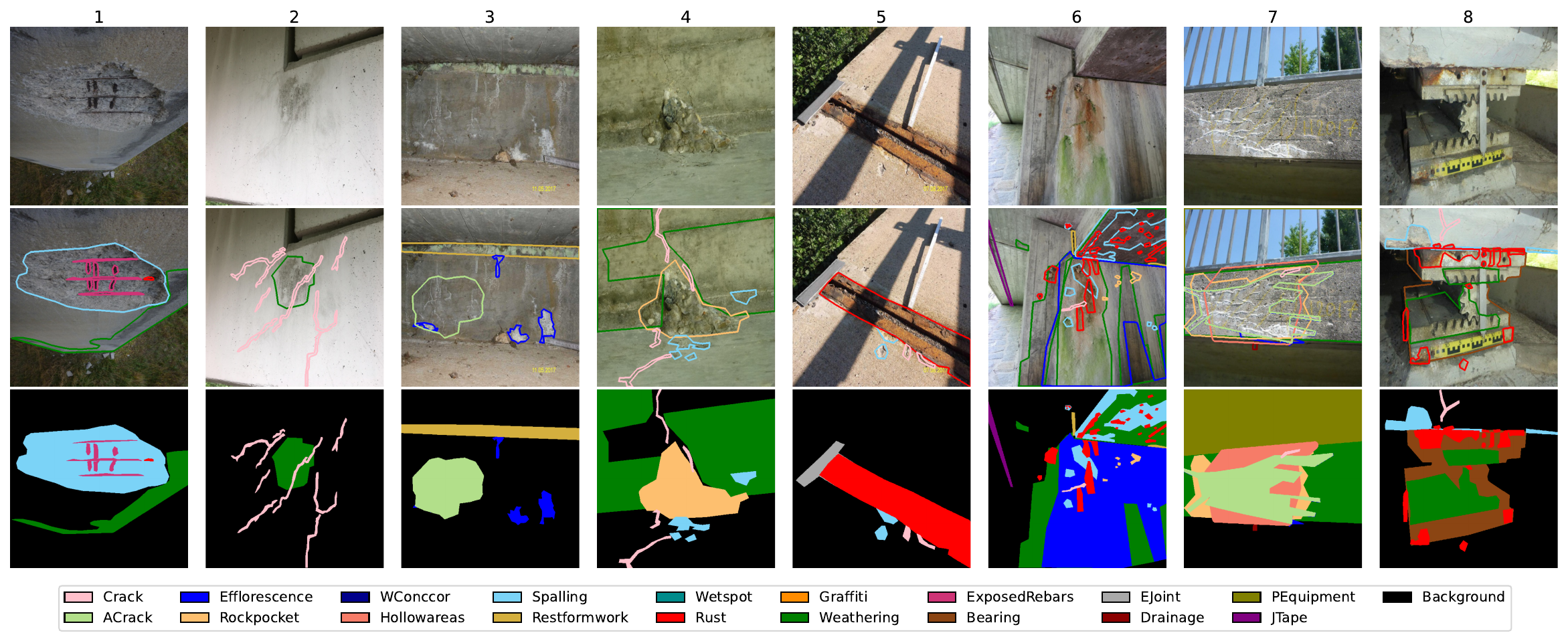}
    \caption{Example annotations from dacl10k. Top row: original image. Middle row: polygonal annotations. Bottom row: stacked masks. The following classes are abbreviated: \textit{Alligator Crack} (ACrack), \textit{Washouts/Concrete corrosion} (WConccor), \textit{Expansion Joint} (EJoint), \textit{Protective Equipment} (PEquipment) and \textit{Joint Tape} (JTape).
    From left to right, the images display the individual classes: 1. \textit{Weathering, Spalling, Exposed Rebars, Rust}; 2. \textit{Weathering, Crack}; 3. \textit{Alligator Crack, Restformwork, Efflorescence}; 4. \textit{Weathering, Crack, Spalling, Rockpocket}; 5. \textit{Crack, Rust, Expansion Joint, Spalling}; 6. \textit{Weathering, Rockpocket, Spalling, Efflorescence, Crack, Rust, Restformwork, Joint Tape}; 7. \textit{Weathering, Protective Equipment, Rockpocket, Efflorescence, Crack, Hollowareas, Alligator Crack, Drainage}; 8. \textit{Weathering, Spalling, Crack, Rust, Bearing}.}
    \label{fig:ExampleAnnots}
    \end{center}
\end{figure*}

Bridges are an essential component of the infrastructure worldwide. They are exposed to many impacts causing damage, such as high traffic loads, extreme weather events, sea salt in coastal areas or treatment with deicing chemicals in cold regions. Due to their age, especially bridges in countries with an economic upswing between the 1950s and 1980s show an increased occurrence of damage by now. These defects are monitored within the scope of bridge inspections aiming to assess the condition of buildings in order to find the ideal timing for rehabilitation steps but also to take immediate actions, \eg, limiting heavy traffic or closing a bridge. Such inspections are usually carried out ``analogously'' by professionally trained civil engineers who visually examine the complete surface of the bridge while taking photos of the defects, and  documenting  damage class, measurements and location on a 2D sketch \cite{AASHTO, RVS, DIN_1076}. 
If these examinations, specifically of buildings in a critical state, could take place more frequently, several bridges may be operated longer and without significant restrictions affecting rail passengers, car drivers and logistics. However, authorities often fail to keep up with the necessary inspection and restoration intervals due to staff shortages and budget limitations, but also because of the commonly applied, time-consuming, analogue inspection process used in practice. 
In many countries, this leads to a steadily growing stock of built structures in poor condition. In governmental reports, the current state of bridges is described as ``1 in 3 U.S. bridges needs repair or replacement ...''~\cite{ARTBA} or ``... at least 25,000 road bridges [in France] are in poor structural condition ...''~\cite{Belin_2022}. This underlines the demand of a more efficient examination pipeline with respect to cost and time \cite{ASCEBridgeReport, SubstandBridgesGreatBrit, Belin_2022, Brueckenstat}. 
The greatest potential for improvement during the bridge inspection process -- irrespective of the used device (unmanned aerial vehicle (UAV) \cite{UAVInspect}, smartphone \cite{SmartphoneBridgeInspec}, augmented reality displays \cite{ARBridgeInspect}) -- is yielded by automating the defect recognition which is crucial to a final assessment in order to determine actions to be taken. The inspection framework that makes use of automated defect recognition is called ``digitized inspection'' (DI). DIs aim for a detailed bridge assessment according to existing country-specific guidelines where the automatized documentation of damage allows reliably classifying, measuring and localizing each existing defect on a given building. Within the scope of DIs, inspectors are strongly supported. The inspections become more efficient and engineers can focus more on the evaluation as well as the context in which defects appear.    

The field of damage recognition on built structures is still unexplored.
In contrast to the fields of autonomous driving \cite{Cordts2016Cityscapes, Geiger2012CVPR, BROSTOW200988, DUS, BDD100K} or medicine \cite{BRATS, simpson2019large, AutoSegChallenge}, semantic segmentation benchmarks for damage recognition are rare. To the best of our knowledge, only two relevant benchmarks in the domain of reinforced concrete defects (RCDs) exist: CrackSeg9k~\cite{CrackSeg9k} and S2DS~\cite{benz2022image}. CrackSeg9k is a collection of various image datasets showing cracked and uncracked surfaces of multiple building materials. 
However, it aims to segment one damage type, whereby, at least 9 types must be recognized in order to be used in practice 
\cite{AASHTO, RVS, DIN_1076, RIEBW}. S2DS is the first multi-class semantic segmentation dataset in RCD domain which includes 743 samples differentiating between five common defects occurring on concrete bridges and control points for georeferencing. Thus, S2DS represents a small variety and complexity with respect to real-world scenarios, with labels assigned to each pixel in a manually exclusive way.

\textit{In conclusion, the significant deterioration of bridges worldwide as well as the lack of visual data for effectively monitoring their defects in a digitized manner emphasize the urgent need for establishing a benchmark in the bridge inspection domain.}
 
We take the problem of semantic segmentation of bridge defects out of the niche by introducing \textit{dacl10k}, the biggest real-world inspection dataset for multi-label semantic segmentation making it possible to perform damage classification, measurement and localization on a pixel-level. Thereby, we enable recognizing 12 frequently occurring defects on reinforced concrete bridges (\eg, \textit{Crack}, \textit{Spalling}, \textit{Efflorescence}) and 6 important building parts (\eg, exposed reinforcement bar \textit{Exposed Rebar}, \textit{Bearing}, \textit{Expansion Joint}, \textit{Protective Equipment}). All these classes play an important role for determining the building's structural integrity, traffic safety and durability. dacl10k includes 9,920 images from more than 100 different bridges, specifically designed for practical use, as it comprises all visually unique damage types defined by bridge inspection standards. dacl10k surpasses previous work significantly in terms of its scale, class variety, and the complex nature of its captured scenes. Besides, we provide essential background knowledge from the civil engineering perspective, which is important for a deeper understanding of the RCD domain.  
In addition, we supply strong baselines to benchmark against. In our model analysis, two semantic segmentation architectures in combination with three encoders are examined.  
The dacl10k dataset, and according baselines, will be publicly released, fostering research within the field of damage recognition on concrete structures using computer vision. 

\section{Related datasets and baselines} 
\label{sec:RelWork}

Within the last six years, major contributions in the field of damage classification on built structures have been made through the introduction of datasets for binary classification \cite{Dorafshan2018Dec,Huthwohl2018Aug,Xu2019,Li2019}, multi-class classification \cite{Huthwohl2019, Bianchi2021}, multi-label classification \cite{Mundt_2019_CVPR}, object detection \cite{Mundt_2019_CVPR}, and semantic segmentation \cite{benz2022image, UAV75, CrackSeg9k}. In the following, we discuss datasets for the last three named tasks. Examples for the subsequently named damage types can be obtained from Figure~\ref{fig:ExampleAnnots}.

Mundt~\etal~\cite{Mundt_2019_CVPR} developed CODEBRIM which is currently the biggest and most realistic dataset for the multi-label classification of RCDs. They differ between the damage types: crack, spallation, exposed reinforcement bar, efflorescence, corrosion and background. 
The unbalanced version of CODEBRIM comprises 7,729 patches of defect images gathered from 30 bridges, chosen based on varying levels of deterioration, defect size, severity, and surface appearance. The images were acquired under changing weather conditions using multiple cameras at varying scales, with high-resolution. A subset of the data was acquired using an UAV, due to the inaccessibility of defects at high locations. Their annotation process was structured as follows: (\textit{i}) selecting bounding boxes (patches) enclosing defects, (\textit{ii}) iterating over the bounding boxes for each damage class and label accordingly, and finally (\textit{iii}) sampling patches of healthy concrete surfaces as well as irrelevant content (background).

For solving the task of binary crack segmentation, Kulkarni~\etal~\cite{CrackSeg9k} combine previously available datasets, inter alia, from the RCD domain. They compile a semantic segmentation dataset, called CrackSeg9k, with 9,255 images of cracks from ten sub datasets on different surfaces. 
Before unifying the datasets, their individual problems (e.g. noise and distortion) are addressed by applying image processing. In addition, they provide baselines where the best model, based on DeepLabv3~\cite{chen2018encoderdecoder}, achieves 77\% mean intersection-over-union (IoU). 

Benz \& Rodehorst~\cite{benz2022image} introduced the Structural Defect Dataset (S2DS) which is the first RCD dataset enabling semantic segmentation of multiple damage types, such as crack, spalling, corrosion, efflorescence, vegetation, and control point which is used for georeferencing. The dataset consists of 743 patches of size 1024x1024 pixel extracted from 8,435 images taken during structural inspections. They used DSLR cameras, mobile phones, and UAVs for acquiring the data. The labeling was executed by one trained computer scientist and had a high level of fineness. Their best model, based on hierarchical multi-scale attention \cite{tao2020hierarchical}, achieves a mean IoU of 92\%, at joint scales of 0.25, 0.5, and 1.0. 

\section{dacl10k dataset}
\label{sec:dacl10k}

dacl10k is the first large-scale dataset for semantic bridge damage segmentation, comprising 9,920 annotated images from real-world inspections. During its creation, our primary objective was to develop a dataset that enables the training of models which later support the inspector during damage recognition and documentation to a maximum. Hence, we analyzed several guidelines determining the level of detail of structural inspections \cite{AASHTO, RVS, DIN_1076, RIEBW}, specifically the visually recognizable defects which must be collected in order to produce a legal bridge assessment. We listed all defects defined by the guidelines accordingly and crossed out the ones that are doppelgangers with respect to visual appearance. Finally, this resulted in the underlying class variety of dacl10k. In the following, we discuss dacl10k's data acquisition, classes, statistics and a comparison to related open-source data.

\subsection{Data acquisition}
Approximately one half of the images originate from databases of engineering offices, while the other half was provided by local authorities from Germany. The images were taken between 2000 and 2020.
Both data sources supplied highly heterogeneous images regarding camera type, pose, lighting condition, and resolution. However, models performing well on dacl10k, most likely generalize well in real-world scenarios.    

\subsection{Damage types and annotation}
\label{sec:DamageTypesAndAnnot}
The 18 classes considered within dacl10k are separated into three groups: concrete defects, general defects and objects. The class names are shown in the first column of Table~\ref{tab:OverallStats}. The concrete defects appear only on building parts made of (reinforced) concrete, while general defects may be present on all materials (\eg, concrete or steel). The only defect within dacl10k that is not visually recognizable, per se, is \textit{Hollowareas}. This damage is usually identified by hammering on the concrete surface, thus, it can only be detected acoustically (not visually) but, as it is bordered with chalk during hands-on inspections, we annotated its markings. 
The objects group includes all components of a bridge that are not made of concrete, such as \textit{Joint Tapes}, Railings or impact attenuation devices (\textit{Protective Equipment}). The objects often show defects such as geometrical irregularities or deficits in structural capacity. Geometrical irregularities can arise from wrong distances between the railing rods or if the railing height is less than the minimum according to the given national standard.   
These visually challenging recognizable issues are not part of the dataset. 
We provide a detailed overview, descriptions and examples of the defect types and objects in the supplementary material.
            
According to the definition of Cordts~\etal~\cite{Cordts2016Cityscapes} our dataset comprises coarse pixel-level annotations. 
We border each defect and object on a given image with one polygon (shape) and assign its label. Furthermore, we include polygons of the same class that overlap with each other in one shape. With respect to inspection standards and application, for all the defect classes, it is not important to differentiate between instances of a given class. Instead, their size and localization on a class-basis is important. 
Thereby, we utilize the open-source labeling tool LabelMe~\cite{Russell2007, LabelMeGithub}. Example annotations are shown in Figure~\ref{fig:ExampleAnnots}. It often appears that shapes of different damage or object classes overlap with each other, \eg, \textit{Spalling} with \textit{Exposed Rebars} covered by \textit{Rust} (see example 1 in Figure~\ref{fig:ExampleAnnots}). Consequently, the underlying task can be described as multi-label semantic segmentation because one pixel can be part of multiple defects and objects. In other words, the labels are not assigned mutually exclusive to the pixels. 

The two main components during labeling are the class guideline and the annotation guideline. The class guideline clearly defines the visual appearance, most commonly occurrence and cause of each defect, \eg, \textit{Efflorescences} (see example 7 in Figure~\ref{fig:ExampleAnnots}):
\begin{itemize}[topsep=1pt,itemsep=1pt,partopsep=1pt, parsep=1pt]
    \item look like stalactites of white to yellowish or reddish color hanging from the bottom of building parts which may also appear to be printed on the buildings surface,
    \item often occur in wet (\textit{Wetspot}) or weathered areas (\textit{Weathering}) of the building and in combination with \textit{Crack} and/or \textit{Rust},
    \item result from the dissolving of salts from the concrete which consequently carbonate.
\end{itemize}
Thus, the annotator -- independently from his domain expertise -- can understand the color, shape and texture of a given class. Furthermore, flags are defined to mark images with personal data (faces, license plates) or images of bad quality. In addition, the provided annotation guideline describes the fineness after which the polygon points shall be set. The goal is to have consistent class and object-distance-dependent density of points over the whole dataset.

Our labeling process consisted of two consecutive steps: 
Annotation of data received from engineering offices by civil engineering students (in-house) and annotation data from the authorities by an external annotation team, previously filtered for relevant image content.  
The students labeled approximately 7,000 images in accordance to our guidelines. Nearly 30\% of the images had to be rejected due to flags indicating bad quality (blurring, overexposure) or personal data. 
The pipeline during in-house labeling was separated into three parts. The first part consisted of the regular annotation which comprises annotating a batch of 100 images, getting feedback by a domain expert and correcting the failures accordingly. The second part included an extensive analysis of the dataset to find structural failures in the annotations. Thirdly, the dataset was divided into subtasks with respect to the failures most commonly made. Then, one student corrected each failure type consecutively. 

The quality assessment of the data annotated by the external team was divided into four quality checks for each data batch. In average, one batch included 250 images. Each check included one iteration over the annotated data by experts. Based on the analysis, the error rate was determined which is the ratio of false-labeled images and total amount of frames in the according batch. Starting with an error rate of 60\%, the rate could be lowered to a final value of 1\%.    

\subsection{Statistical analysis}
\label{sec:StatAnal}

\begin{figure}
    \includegraphics[width=\linewidth]{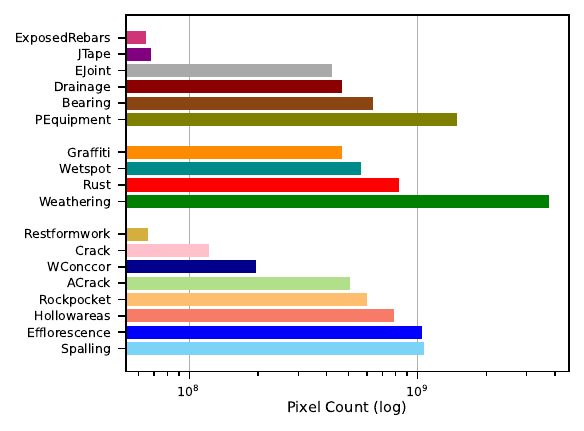}
    \caption{Pixel counts with respect to each class in dacl10k based on the original image sizes. The bars are arranged according to the group affiliation.}
    \label{fig:bar_chart}
\end{figure}
In the following, we discuss split-independently (\textit{i}) the classwise pixel counts, (\textit{ii}) averages and shares regarding the polygons and pixels as well as (\textit{iii}) the split-dependent statistics, based on the original image resolutions.  

Regarding the pixel counts over the whole dataset (see Figure~\ref{fig:bar_chart}), it can be stated that \textit{Weathering}, followed by \textit{Protective Equipment}, are the most dominant classes with nearly four and 1.5 billion pixels. Within the range of 0.1 billion and 1 billion pixels, the majority of defects and objects with respect to the number of pixels can be found. Clearly underrepresented are the defects \textit{Restformwork} (66 million) and the objects \textit{Joint Tape} (68 million) and \textit{Exposed Rebars} (65 million). 

The average image size is 1581 px in height and 1950 px in width. The mean image area is approximately 4 megapixels while the total pixel area in the dataset is approx. 43 billion px. 
Table~\ref{tab:OverallStats} provides an overview of statistics describing the classwise density of polygons, size of polygons, density of pixels, share of polygons and share of pixels over the whole dataset. 
In average, 1.8 crack shapes are present on a crack image, whereby, one polygon includes 27,467 px. The average crack image shows approx. 50,000 px labeled as crack. 
With respect to the displayed shares, we observe that four out of 100 polygons are labeled as crack and 0.3\% of the total pixel area received the label \textit{Crack}.
Furthermore, Table~\ref{tab:OverallStats} reveals the cause of the overrepresented classes \textit{Weathering} and \textit{Protective Equipment}. They display a share regarding the number of polygons of 5.31\% (top 20\%) and 2.09\% (exactly the median). 
This, in combination with the fact that an according image shows 900,000 px or rather 786,000 px of that class, leads to their dominant role. The overrepresentation of \textit{Weathering} can be more fatal than the one of \textit{Protective Equipment} with respect to the model performance.  
This is due to the fact that the features (shape and texture) of \textit{Weathering} are similar to the ones from \textit{Wetspot}. Both are of round shape and represented by a darker area surrounded by a brighter ``rest''. They vary slightly with regards to their texture. \textit{Weathering} is more noisy and more matt than \textit{Wetspot} which is smooth and sometimes mirroring. 
In addition, \textit{Wetspot} and \textit{Weathering} often overlap which makes it difficult to distinguish between them from the model's perspective.
The features of the object \textit{Protective Equipment}, in contrast, are unique and therefore shouldn't interfere with other classes during learning. 
The lack of pixels representing \textit{Restformwork}, \textit{Exposed Rebars} and \textit{Joint Tape}, as mentioned before, originates from their relatively rare occurrences but mostly from their small shapes. In average, polygons bordering \textit{Restformwork} or \textit{Joint Tape} have a size of approx. 50,000 px while polygons labeled as \textit{Exposed Rebars} include 26,000 px. The average size of their polygons is equal to or less than the lower quartile (50,367 px). Additionally, \textit{Exposed Rebars} shows the smallest average polygon size of all classes.          

\begin{table}
\begin{center}
\scriptsize      
\begin{tabular}{L{13.5mm}R{9mm}R{9mm}R{10mm}R{8.5mm}R{8mm}}
\toprule
Class        & \#polyg./ image & \#pixels/ polyg. & \#pixels/ image & \%polyg. & \%pixels \\ \midrule
Crack         & 1.81             & 27,467            & 49,605          & 4.02       & 0.30      \\
ACrack        & 1.12             & 950,694           & 1,064,777        & 0.48       & 1.25     \\
Efflorescence & 2.29             & 208,565           & 478,395         & 4.54       & 2.59     \\
Rockpocket    & 4.75             & 50,564            & 240,079         & 10.74      & 1.48     \\
WashoutsC.     & 1.34             & 807,721           & 1,079,936        & 0.22       & 0.48     \\
Hollowareas   & 1.21             & 415,627           & 504,482         & 1.72       & 1.96     \\
Spalling      & 2.68             & 83,298            & 223,185         & 11.60       & 2.64     \\
Restformw.  & 1.19             & 50,170            & 59,757          & 1.18       & 0.16     \\ \midrule
Wetspot       & 1.48             & 271,408           & 400,778         & 1.89       & 1.40      \\
Rust          & 3.62             & 46,680            & 168,997         & 16.01      & 2.04     \\
Graffiti      & 2.29             & 172,917           & 395,596         & 2.44       & 1.15     \\
Weathering    & 1.41             & 639,776           & 903,974         & 5.31       & 9.28     \\ \midrule
ExposedR. & 2.25             & 25,770            & 58,034          & 2.26       & 0.16     \\
Bearing       & 1.45             & 421,784           & 612,581         & 1.37       & 1.57     \\
EJoint        & 1.12             & 700,054           & 783,023         & 0.55       & 1.05     \\
Drainage      & 1.37             & 230,815           & 316,099         & 1.83       & 1.15     \\
PEquipment    & 1.22             & 641,715           & 785,601         & 2.09       & 3.67     \\
JTape         & 1.19             & 49,401            & 58,569          & 1.25       & 0.17     \\ \midrule
Background    & 3.47             & 871,631           & 3,020,476        & 30.51      & 72.69    \\
\bottomrule
\end{tabular}
\caption{Overall statistics of the dataset regarding average number of polygons per image, number of pixels per polygon, number of pixels per image,  share of polygons and share of pixels. Midrules separate the classes according to their group affiliation.}
\label{tab:OverallStats}
\end{center}
\end{table}

To ensure similar data distributions (regarding damage and object classes) in each split, we create data partitions according to image similarity. For that, we employ K-means clustering~\cite{KMeans} with 20 clusters. Then, we proportionally draw samples from each cluster for the train (70\%), valid (10\%), testdev (10\%), and testpriv (10\%) set with respect to the number of polygons and images (see Table~\ref{tab:SplitStats}). The testdev and testpriv split are summarized within the test split. 

The Table presents the classwise statistics of dacl10k separated into the given splits. Again, taking the \textit{Crack} class as an example, the aforementioned proportions can be observed in terms of number of pixels with a number of 89,316,599 px (73\%) in the train and approx. 11,000 px (9\%) in the validation split. 
Compared to the validation split, the values of the test split are approx. twice as high.
For the number of polygons and images, Table~\ref{tab:SplitStats} displays the same targeted proportions as for the pixel count. 

\begin{table*}
\begin{center}
\footnotesize
\begin{tabular}{lrrrrrrrrr}
\toprule
\multirow{2}{*}{Class} & \multicolumn{3}{c}{Train}                                       & \multicolumn{3}{c}{Valid}                                       & \multicolumn{3}{c}{Test}                                                \\   \cmidrule(lr){2-4} \cmidrule(lr){5-7} \cmidrule(lr){8-10} 
                        & \#pixels    & \#polyg. & \#images & \#pixels    & \#polyg. & \#images & \#pixels    & \#polyg. & \#images  \\ \midrule 
Crack         & 89,316,599          & 3,092               & 1,720               & 11,462,336          & 457                 & 254                 & 21,199,921                & 892                       & 485                       \\
ACrack        & 379,337,903         & 378                 & 336                 & 33,145,894          & 48                  & 42                  & 93,285,498                & 106                       & 97                        \\
Efflorescence & 773,838,619         & 3,378               & 1,523               & 69,295,099          & 502                 & 206                 & 204,072,743               & 1,141                     & 460                       \\
Rockpocket    & 416,831,564         & 8,241               & 1,712               & 51,701,480          & 1,207               & 259                 & 131,663,248               & 2,422                     & 529                       \\
WConccor      & 153,351,739         & 176                 & 133                 & 7,781,177           & 23                  & 15                  & 34,335,547                & 43                        & 33                        \\
Hollowareas   & 589,108,842         & 1,327               & 1,100               & 68,277,162          & 193                 & 155                 & 133,137,140               & 382                       & 312                       \\
Spalling      & 754,421,298         & 8,638               & 3,289               & 94,740,019          & 1,444               & 485                 & 218,557,811               & 2,736                     & 1,010                     \\
Restformwork  & 42,413,460          & 841                 & 716                 & 6,143,333           & 164                 & 132                 & 17,116,171                & 304                       & 251                       \\ \midrule
Wetspot       & 402,563,879         & 1,436               & 972                 & 47,002,103          & 216                 & 144                 & 117,133,455               & 436                       & 298                       \\
Rust          & 592,984,180         & 12,272              & 3,451               & 79,135,671          & 1,801               & 465                 & 153,938,160               & 3,623                     & 972                       \\
Graffiti      & 308,779,763         & 1,866               & 797                 & 71,491,732          & 317                 & 146                 & 85,741,013                & 512                       & 235                       \\
Weathering    & 2,563,181,494       & 4,056               & 2,830               & 367,950,798         & 572                 & 407                 & 823,072,419               & 1,240                     & 916                       \\ \midrule
ExposedRebars & 44,829,810          & 1,720               & 773                 & 3,824,934           & 244                 & 104                 & 15,821,469                & 538                       & 234                       \\
Bearing       & 431,016,513         & 1,039               & 731                 & 89,236,423          & 160                 & 105                 & 116,219,161               & 310                       & 203                       \\
EJoint        & 335,469,869         & 446                 & 396                 & 21,145,699          & 56                  & 51                  & 66,216,825                & 102                       & 93                        \\
Drainage      & 368,660,666         & 1,393               & 1,030               & 35,297,153          & 244                 & 151                 & 62,288,022                & 383                       & 294                       \\
PEquipment    & 1,070,432,844       & 1,616               & 1,320               & 103,082,479         & 211                 & 175                 & 312,055,649               & 488                       & 396                       \\
JTape         & 47,582,309          & 911                 & 772                 & 8,569,024           & 152                 & 128                 & 12,022,699                & 317                       & 264                       \\ \midrule
Background    & 21,196,542,072      & 23,347              & 6,801               & 2,681,148,040       & 3,279               & 962                 & 5,514,566,563             & 7,095                     & 1,968  \\                
\bottomrule
\end{tabular}
\caption{Splitwise statistics regarding the number of pixels, polygons and images.}
\label{tab:SplitStats}
\end{center}
\end{table*}

\subsection{Comparison to other datasets}
Compared to CrackSeg9k~\cite{CrackSeg9k} and S2DS~\cite{benz2022image}, the crack annotations of dacl10k are coarser. CrackSeg9k is a collection of multiple available binary crack segmentation datasets where each was acquired in a standardized setting with respect to camera pose, lighting condition and hardware. S2DS is a single-label semantic segmentation dataset that includes images of RCDs (and control points) captured during real structural inspections, where the fineness of crack annotations is also high.  
According to the inspection guidelines, the recognition of cracks requires the highest degree of accuracy because their width plays an important role during their assessment. \Eg, in Germany the minimum crack width, which must be documented, is 0.2mm \cite{DIN_1076, RIEBW}. Consequently, for the damage type \textit{Crack}, finer annotations are definitely useful when it comes to practical use.  

To sum up, both related datasets show a higher level of detail regarding crack annotations than dacl10k but are less diverse with respect to class variety and real-world scenarios. CrackSeg9k enables the training of models for binary crack segmentation only. The annotations of S2DS provide single-target information at most, meaning that one pixel can only belong to one class. In addition, S2DS consists of a relatively small number of images. 
However, we focus on the multi-label semantic segmentation of all visually unique defects and objects on concrete bridges. Thereby, we take into account the frequently occurring case that in real-world scenarios multiple defects overlap.    
Regarding the crack annotations, it can be stated that a pixel-accurate classification of our crack data may be possible by applying methods from the field of weakly-labeled data \cite{SegmentEveryThing}. In the CityScapes dataset~\cite{Cordts2016Cityscapes}, for example, the majority of the samples were coarsely annotated (20,000 images) with the intention to foster research specifically in this field. 
With regards to applications within the framework of existing standards, all other classes in our dataset do not require a finer annotation and prediction, respectively. \Eg, during currently practiced analogue inspections, the diameter and corresponding area of a \textit{Spalling} is usually measured in a rough manner with a folding rule which is sufficient for its assessment. 

\section{Baselines}
\label{sec:baselines}
In the following, we describe the development of the baseline models and their test results. In order to evaluate the baselines and to demonstrate challenges with respect to the annotation of the underlying data, we conduct an ``Engineer versus Machine'' (EvsM) comparison. Finally, we discuss the results incorporating the findings regarding dacl10k, the baselines and the EvsM comparison.

\subsection{Implementation details}
For the development of the baselines we analyze two different CNN-based semantic segmentation architectures with three
different encoders, and one Transformer-based model.
We used DeepLabV3+~\cite{chen2017rethinking, chen2018encoderdecoder} and Feature Pyramid Network (FPN)~\cite{kirillov2019panoptic} as CNN baseline architectures which represent powerful models often used in research and industry.
As encoders we utilize MobileNetV3-Large~\cite{howard2019searching} (3M parameters), EfficientNet-B2~\cite{tan2020efficientnet} (7M parameters) and EfficientNet-B4 (17M parameters). 
For each of these base models we also investigate a separate model using an auxiliary loss for multi-label classification. 
The auxiliary classification head placed after the encoder consists of an average global pooling layer, followed by a dropout and linear layer. The weights of the model are updated based on a combined weighted loss including the mask and auxiliary loss. The total loss $\mathcal{L}_{total}$ is computed as follows:
\begin{align}
    \mathcal{L}_{total} = \mathcal{L}_{mask} + 0.1\times \mathcal{L}_{aux}
    \label{eq:loss_total}
\end{align}
where $\mathcal{L}_{mask}$ and $\mathcal{L}_{aux}$ are based on Dice~\cite{Sudre2017dice} and Cross Entropy loss respectively.

To meet current developments in Transformer-based models, we trained a SegFormer model~\cite{xie2021segformer}. Since auxiliary loss is unusual for this network, only Dice loss is utilized.
Adam optimizer \cite{kingma2017adam} with four different learning rates ($5e^{-3}$, $1e^{-3}$, $5e^{-4}$, $1e^{-4}$) is applied whereof the model with the best loss based on the validation split is reported.
We make use of a cosine learning rate scheduler with a warm-up phase over two epochs and train each model for 30 epochs. 
All models are initialized with ImageNet weights~\cite{ImageNet, rw2019timm}.

The images and annotations are resized to a resolution of $512\times512$ before being fed to the network. Thereby, each defect and object class is considered with a separate binary mask, leading to a total number of 18 channels. All following results are reported on the same resolution, enabling an evaluation that is not focused towards large images.

\subsection{Baseline results}
\label{sec:BaseRes}
In order to find the best performing model on dacl10k, we compare the mean IoU of seven different models (see Table~\ref{tab:model-table}).
The best network using the DeepLabV3+ architecture includes an EfficientNet-B4 backbone without considering auxiliary loss. It achieves a mean IoU of 0.411. 
The highest mean IoU is obtained at a value of 0.414 by the model consisting of an EfficientNet-B4 encoder and FPN architecture while taking the auxiliary loss into account. 
For both, FPN and DeepLabV3+, it can be observed that the more parameters the encoder has, the higher is the reached mean IoU. The SegFormer model achieves a mean IoU of 0.400 which is less than the best model based on DeepLabV3+ or FPN.   

\begin{table}[t]
\scriptsize
\centering
\begin{tabular}{@{}ccccccccc@{}}
\toprule
   \multirow{2}{*}{Aux}    & \multicolumn{3}{c}{DeepLabv3+} & \multicolumn{3}{c}{FPN} & \multirow{2}{*}{SegFor}\\ \cmidrule(lr){2-4} \cmidrule(lr){5-7}
   & \multicolumn{1}{l}{MN} & \multicolumn{1}{l}{EN-B2} & \multicolumn{1}{l}{EN-B4} & \multicolumn{1}{l}{MN} & \multicolumn{1}{l}{EN-B2} & \multicolumn{1}{l}{EN-B4} \\ \midrule
 $-$ &  0.320 & 0.360 & 0.411 & 0.376 & 0.384 & 0.364 & 0.400\\
 \Checkmark & 0.329 & 0.400  & 0.409 & 0.378  & 0.395 & \textbf{0.414} & $-$\\
 \bottomrule
\end{tabular}
\caption{Mean IoU on valid split for both architectures and three encoders (MobileNetV3-Large, EfficientNet-B2 and EfficientNet-B4) with and without auxiliary loss, and the SegFormer model.}
\label{tab:model-table}
\end{table}

Table~\ref{tab:testdev-table1} displays the classwise IoUs and the mean IoU of the best model, described in the preceded paragraph. We report results on the validation and test split (see Table~\ref{tab:SplitStats}). 
Due to the small differences between the metrics over the splits, it can be stated that the data, with respect to complexity, is evenly distributed. In the following, the results on the test split are documented. 
The lowest IoU is obtained for the defect \textit{Washouts/Concrete Corrosion} with a value of 0.121. This class is not underrepresented (see Figure~\ref{fig:bar_chart}). We observe that its texture and shape (features) are very familiar to the ones from \textit{Rockpocket} and \textit{Spalling} which both are strongly represented. 
Other classes for which a low IoU is reported are \textit{Wetspot}, \textit{Restformwork}, \textit{Crack} and \textit{Rockpocket}. The possible reason for the bad performance on \textit{Wetspot} is explained in Section~\ref{sec:StatAnal}. \textit{Restformwork} has many different visual appearances, which the model probably fails to summarize within one class. The \textit{Crack} class, in average, is represented by the least number of pixels per image, and it has the smallest polygons, after \textit{Exposed Rebars}, as they are elongated and very narrow (see Table~\ref{tab:OverallStats}). Thus, the IoU is a challenging metric for this defect, as the false-negative segmentation of classes represented by more pixel area is less penalized. 
On \textit{Rockpocket} a low IoU is obtained because of the aforementioned similar-looking defects. 
The best results can be reported for the objects \textit{Protective Equipment} (0.715) and \textit{Bearing} (0.564) while the best general defect is \textit{Graffiti} (0.623) and with respct to concrete defects, a good IoU can be observed for \textit{Hollowareas} (0.555) and \textit{Alligator Crack} (0.482). 
\textit{Graffiti} and \textit{Protective Equipment} are the two most individual classes regarding their visual appearance, additionally, they are well represented (see Figure~\ref{fig:bar_chart}) which explains their high IoU compared to the rest of the classes. Summarizing, the best model achieves a mean IoU of 0.424. A more detailed analysis of the problematic classes is provided within the supplementary material.  

\begin{table}[ht]
\begin{center}
\small
\begin{tabular}{lrr}
\toprule
Class                & valid &  test \\ \midrule
Crack                & 0.288 &  0.286 \\
ACrack               & 0.473 &  0.482 \\
Efflorescence        & 0.338 &  0.415 \\
Rockpocket           & 0.267 &  0.294 \\
WConccor             & 0.085 &  0.121 \\
Hollowareas          & 0.536 &  0.555 \\
Spalling             & 0.374 &  0.406 \\
Restformwork         & 0.336 &  0.285 \\\midrule
Wetspot              & 0.232 &  0.243 \\
Rust                 & 0.414 &  0.450 \\
Graffiti             & 0.586 &  0.623 \\
Weathering           & 0.423 &  0.395 \\\midrule
ExposedRebars        & 0.393 &  0.358 \\
Bearing              & 0.676 &  0.564 \\
EJoint               & 0.474 &  0.524 \\
Drainage             & 0.521 &  0.563 \\
PEquipment           & 0.675 &  0.715 \\
JTape                & 0.362 &  0.362 \\\midrule
Mean                 & 0.414 &  0.424 \\
\bottomrule
\end{tabular}
\caption{Classwise and mean IoU of the best model (FPN with EfficientNet-B4 and auxiliary loss) on the validation and test split.}
\label{tab:testdev-table1}
\end{center}
\end{table}

\section{Engineer vs. Machine}
\label{sec:engvsmachine}

\begin{figure*}[ht]
\begin{center}
    \includegraphics[width=1\textwidth]{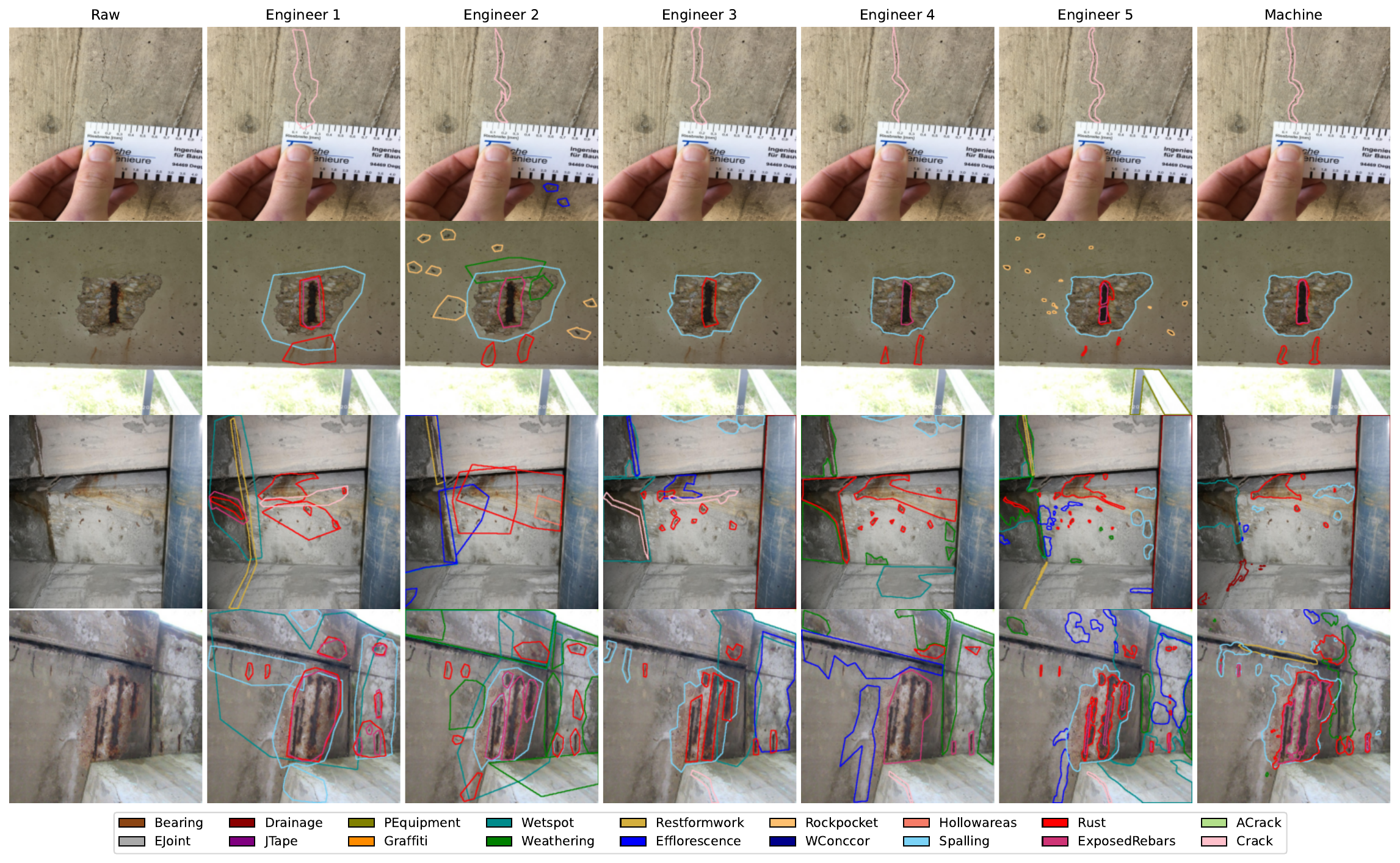}
    \caption{Qualitative evaluation of the best model on four samples (Raw) from the testdev split by comparing annotations of five civil engineers (Engineer 1-5) with the best model's predictions (Machine). From top to bottom, the samples show: \textit{Crack}; a few tiny cavities which are considered as \textit{Rockpockets} and \textit{Spalling, Exposed Rebars, Rust}; \textit{Wetspot, Weathering, Efflorescence, Crack, Spalling, Exposed Rebars and Rust}; \textit{Wetspot, Weathering, Efflorescence, Restformwork, Drainage, Rust, Spalling}.}
\label{fig:EngVsMachine} 
\end{center}
\end{figure*}

Figure~\ref{fig:EngVsMachine} displays the EvsM comparison, which enables a qualitative validation of segmentations from the best model (see Section~\ref{sec:BaseRes}) by comparing the annotations of civil engineers with the network's predictions. Therefore, we asked the five experts, two of whom perform bridge inspections frequently, to annotate four representative samples drawn from dacl10k's testdev split. Two samples are considered easy and the others difficult to evaluate because of their low resolution, diversity regarding the lighting condition, camera pose as well as defect types and objects. After an introduction to the class guideline, the experts were instructed to annotate the images with the same quality they would expect an application to highlight defects during a close-up or hands-on bridge inspection. Compared to other inspection types, hands-on inspections require the highest quality with respect to defect classification, measurement and localization, which also includes the detection of \textit{Hollowareas} by hammering the concrete surface.   
Regarding the differences among engineers, it can be stated that with increasing complexity of the samples also the variance of the chosen labels rises. Engineer 5 annotated at the highest quality level. Thus, especially his annotation is used as a qualitative benchmark for our baselines prediction. 
For the first two samples, which contain fewer classes compared to the last two, the prediction of the machine is better than the average annotation of the five engineers.
Only in the second image the small cavities that are part of the defect \textit{Rockpocket} and the \textit{Protective Equipment}, which is overexposed, are not recognized. The last two images reveal the limits of our baseline. 
For the third sample the classes \textit{Wetspot}, \textit{Rust} and \textit{Drainage} are well predicted, whereas, areas of \textit{Efflorescence} and weak \textit{Spallings} are not segmented. \textit{Weathering} is not present at all in the model's prediction.  
On the last image the model doesn't predict the defects \textit{Crack} at the center bottom and the \textit{Wetspot} which is located mainly on the wall. Furthermore, the model classifies \textit{Restformwork} which is not present on the image, while \textit{Spalling} and \textit{Exposed Rebars} are correctly recognized. The remaining classes are partially classified.

\section{Discussion}
\label{sec:discussion}

We have introduced the first large-scale dataset and corresponding baselines for multi-label semantic segmentation in the bridge inspection domain. It's based on real-world data, with a label distribution deriving from the visually distinguishable classes of multiple country-specific inspection standards. It is important to note that the concrete and general defect group labels are not restricted to bridges, as they can occur on any building made of (reinforced) concrete. 
The evaluation of the baselines, especially the EvsM comparison, generally shows a good performance. However, limitations can be observed for classes with minimal feature differences between each other, \eg, \textit{Weathering} and \textit{Wetspot} or \textit{Washouts/Concrete Corrosion}, \textit{Rockpocket} and \textit{Spalling}. In addition, the datatset is highly unbalanced, leading to biases towards the overrepresented classes.  

We are confident that a more sophisticated search for architectures and hyperparameters as well as data augmentation methods will lead to substantial improvements. Additionally, with respect to finer \textit{Crack} segmentations, approaches from the field of weakly-labeled data may better satisfy the geometrical accuracy requirements of this defect type. 
Overall, we believe that due to its size and diversity, dacl10k makes an important contribution to the field of automated structural inspection.

\noindent\textbf{Acknowledgements.} We thank Johannes Kreutz, Helmut Mayer and Wolfgang Kusterle for their insights and suggestions. We are most grateful to the engineering offices and authorities for the supply of inspection images. We also thank the institute for Distributed Intelligent Systems at the University of the Bundeswehr Munich for providing computing time at MonacumOne Cluster. The project was funded by the Bavarian Ministry of Economic Affairs (MoBaP research project, IUK-1911-0004// IUK639/003). 

\clearpage

{\small
\bibliographystyle{ieee_fullname}
\bibliography{egbib}
}

\clearpage

\appendix
\section{dacl10k}
\subsection{Additional statistics}
Table~\ref{tab:AddStats} displays additional statistics of dacl10k. The average image has a picture format of 4:3 and comprises four megapixels. In total, dacl10k includes 40 billion pixels and 110,533 polygons. The average number of polygons per image amounts to eleven. 

\begin{table}[ht]
    \centering
    \begin{tabular}{lr}
         \toprule
         \#images & 9,920\\
         Image width (min, mean, max) & 336, 1950, 6000 \\
         Image height (min, mean, max) &  245, 1581, 5152\\
         \#pixels of image areas & 40,435,268,789 \\ 
         Average \#pixels/image & 4,076,136 \\
         Average \#polygons/image & 11 \\
         \bottomrule
    \end{tabular}
    \caption{Additional statistics on dacl10k.}
    \label{tab:AddStats}
\end{table}

\subsection{Accessibility}
The dacl10k dataset is made freely available to academic and non-academic entities for non-commercial purposes such as academic research, teaching, scientific publications, or personal experimentation. Permission is granted to use the data, given that you agree to the Attribution-NonCommercial 4.0 International (CC BY-NC 4.0) license. 

\subsection{Class descriptions}
In Table~\ref{tab:ConcreteDefects}~(Concrete Defects), Table~\ref{tab:GeneralDefects}~(General Defects) and Table~\ref{tab:Objects}~(Objects) a detailed description and an example image for each class are listed. Within these tables, we describe the visual appearance and the cause of the defect, or rather the functionality of the given object. The examples include the annotations of the according class exclusively. Additionally, the class abbreviations -- used in the tables and figures of this work as well as in the annotation files of dacl10k -- are given in parentheses.  

For a deeper understanding of the concrete defects it is important to note that ordinary concrete consists of cement, water, sand and coarse aggregate (gravel). In its unhardened state (wet concrete) the mix of cement and water is named cement paste, while the hardened paste is called cement stone. The cement paste -- later cement stone -- binds the other concrete components, sand and gravel, together.
Regarding the concrete defects \textit{Crack}, \textit{Alligator Crack} and \textit{Spalling} it is noteworthy that concrete has a high compressive strength but a low tensile strength ($\approx$10\% of its compressive strength). The overstressing of its tensile strength leads to the occurring of the according defect. The cause of the overstressing differs for each of these defects, which is further explained in Table~\ref{tab:ConcreteDefects}.  

\subsection{Annotation-specific problems}
Only a few related datasets are available (see main paper), which are limited to five RCDs. In addition, the labeling of dacl10k requires deep domain knowledge. Furthermore, dacl10k was labeled by two different groups of annotators, civil engineering students and a professional annotation team. On the one hand, civil engineers have in-depth knowledge about RCDs and bridge components. On the other hand, they are usually not familiar with computer vision problems and, therefore, are not aware of the caveats during the development of semantic segmentation datasets. The professional annotation team, against it, has no domain expertise in regards to bridge defects and objects. Thus, their understanding of important context is less pronounced. 

\subsection{Data-specific problems}
\begin{figure*}[ht]
\begin{center}

\begin{subfigure}[b]{.85\textwidth}
    \includegraphics[height=.33\textwidth]{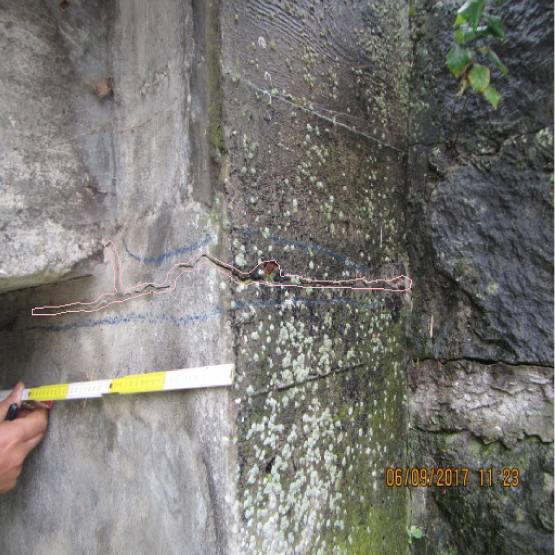}\hfill
    \includegraphics[height=.33\textwidth]{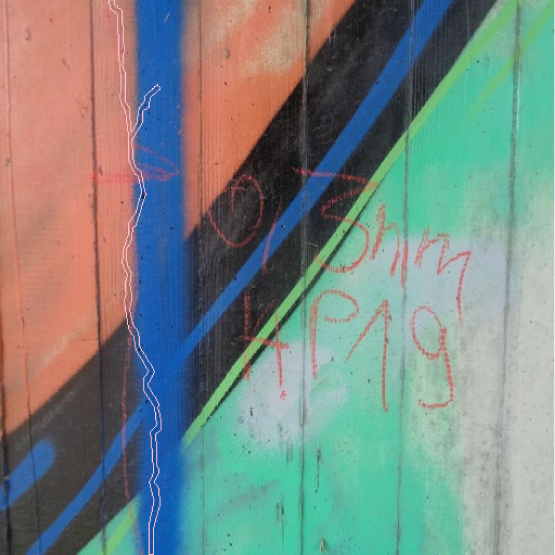}
    \includegraphics[height=.33\textwidth]{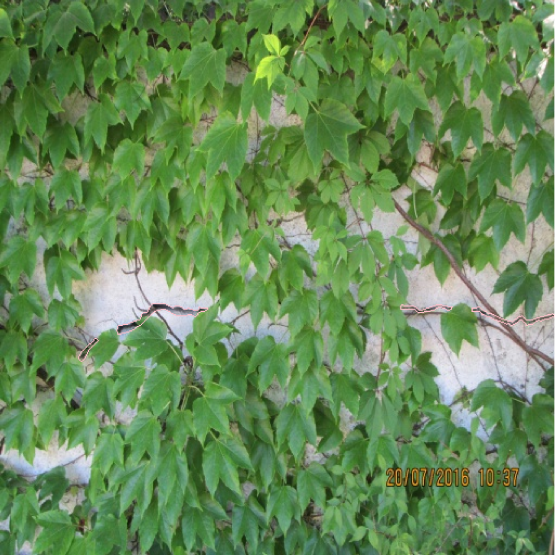}\hfill
 \caption{Challenging samples showing a \textit{Crack}: in combination with strong \textit{Weathering}, in combination with \textit{Graffiti} and partially overlapped by vegetation (\textit{Weathering}).}
   \label{fig:DiffCrack} 
\end{subfigure}

\begin{subfigure}[b]{.85\textwidth}
    \includegraphics[height=.33\textwidth]{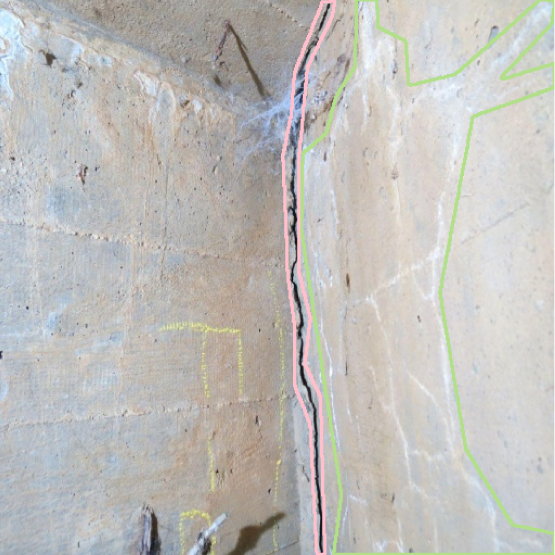}\hfill
   \includegraphics[height=.33\textwidth]{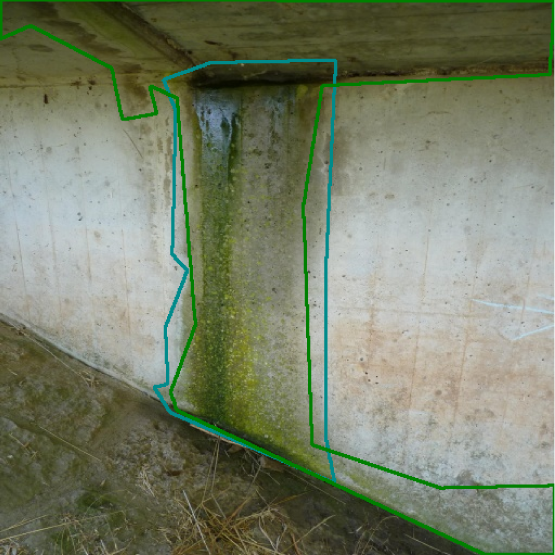}\hfill
    \includegraphics[height=.33\textwidth]{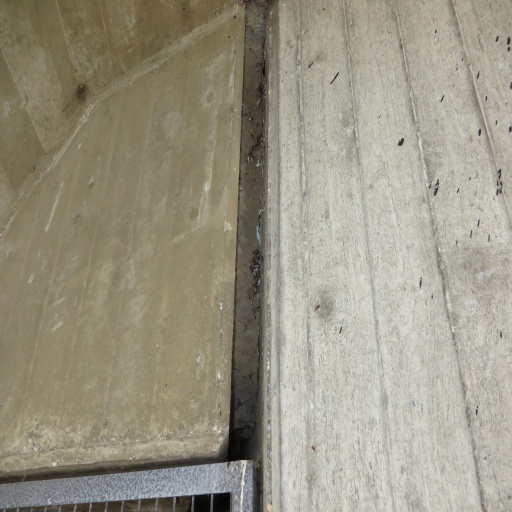}
 \caption{\textit{Crack} vs. \textit{Alligator Crack}, \textit{Weathering} vs. \textit{Wetspot}, \textit{Restformwork} covered with concrete slurry (no ground-truth annotation for better visibility).}
   \label{fig:BadPerformance} 
\end{subfigure}

\begin{subfigure}[b]{.85\textwidth}
    \includegraphics[height=.33\textwidth]{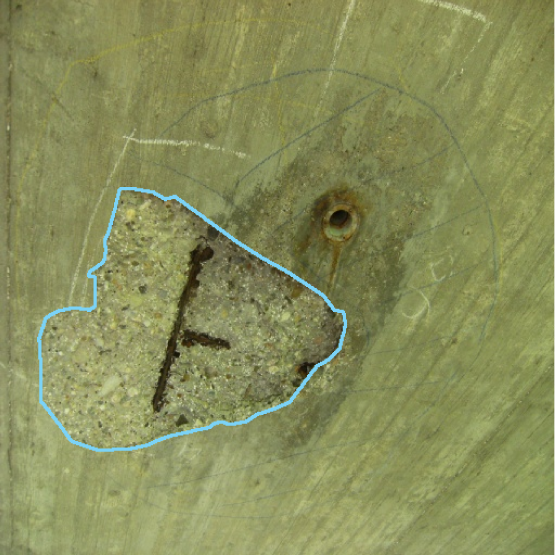}\hfill
    \includegraphics[height=.33\textwidth]{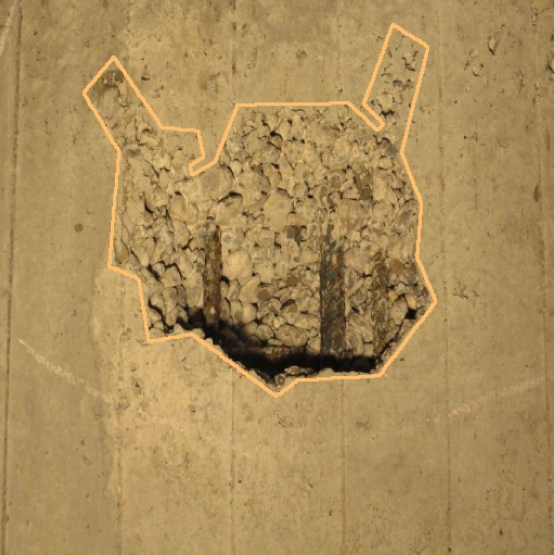}\hfill
   \includegraphics[height=.33\textwidth]{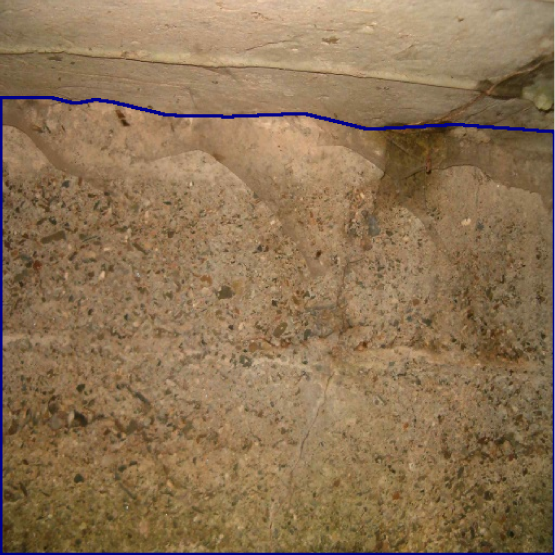}
 \caption{\textit{Spalling}, \textit{Rockpocket}, \textit{Washouts/Concrete corrosion}.}
   \label{fig:EasyMixedUp} 
\end{subfigure}
\caption{Challenging defects. All subcaptions describe the images from left to right.}
\label{fig:ProblemClasses} 
\end{center}
\end{figure*}

\begin{figure*}[ht]
\begin{center}
\begin{subfigure}[b]{.85\textwidth}
    \includegraphics[height=.33\textwidth]{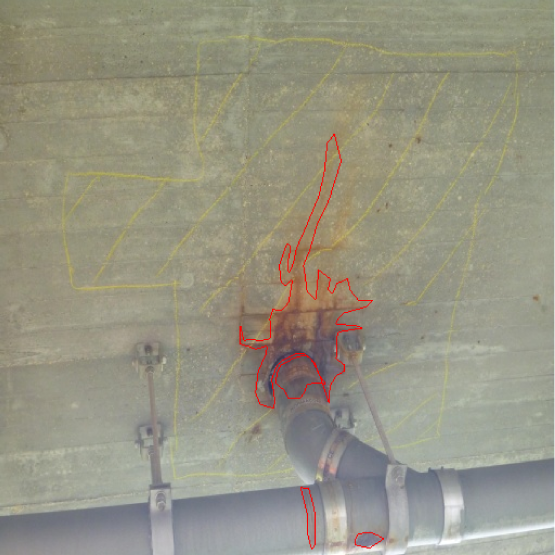}\hfill
    \includegraphics[height=.33\textwidth]{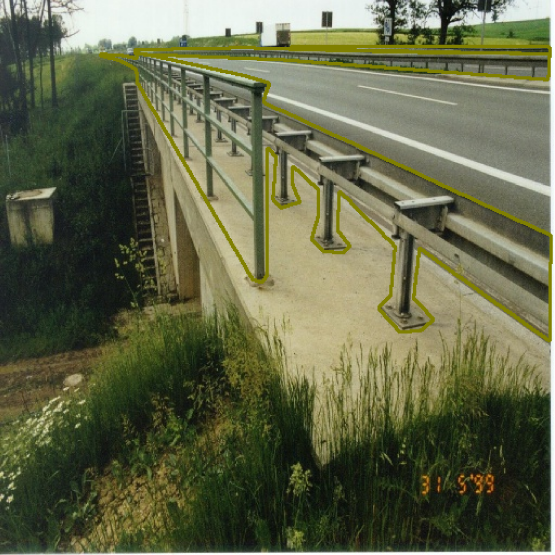}\hfill
    \includegraphics[height=.33\textwidth]{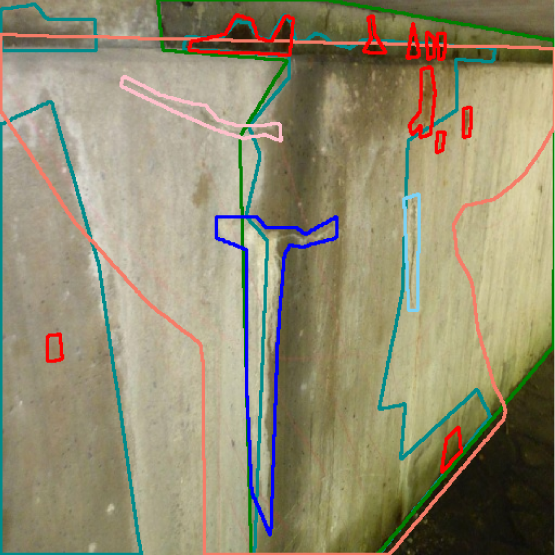}
     \caption{No clear border of the defect (\textit{Rust}), the class (\textit{Protective Equipment}) is stretched from fore- to background.Many overlapping classes with partially low contrast like the border and hatching of the \textit{Hollowarea} (pale red), and the \textit{Efflorescence} (dark blue) next to the bright, dry and healthy concrete surface.}
   \label{fig:ClassProblems} 
\end{subfigure}
\begin{subfigure}[b]{.85\textwidth}
   \includegraphics[height=.33\textwidth]{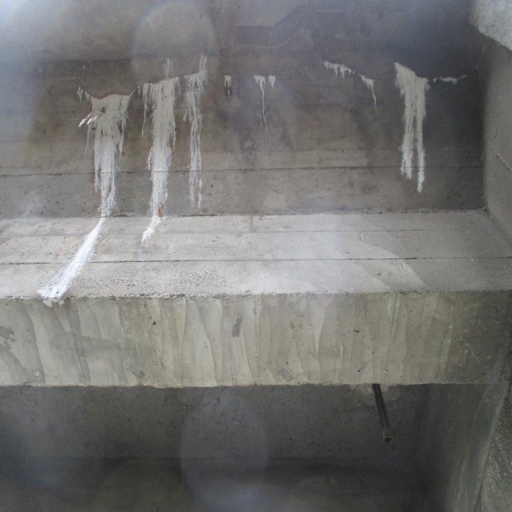}\hfill
   \includegraphics[height=.33\textwidth]{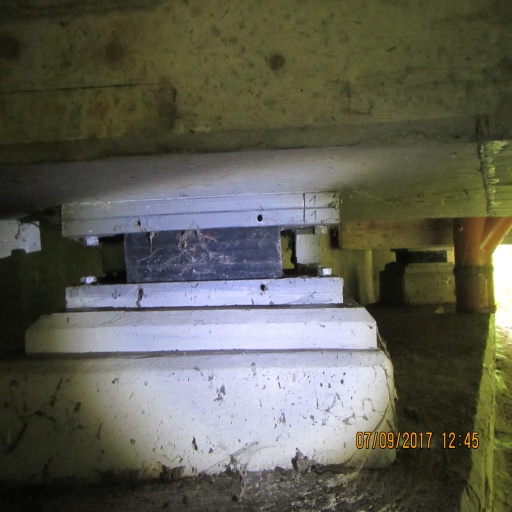}\hfill
   \includegraphics[height=.33\textwidth]{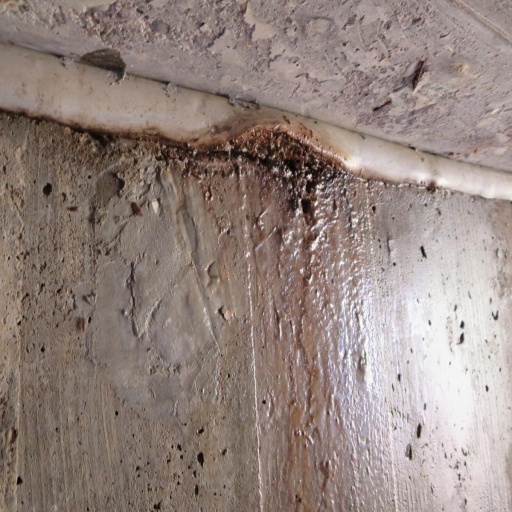}
   \caption{Image quality problems: lens flares, partial overexposures due to flashlight, and sunlight.}
   \label{fig:ImgQualProb} 
\end{subfigure}
\caption{General problems across the dataset. All subcaptions describe the images from left to right.}
\label{fig:GeneralProblems}
\end{center}
\end{figure*}

In the following, we discuss the most challenging defects (see Figure~\ref{fig:ProblemClasses}) and general problems across the dataset (see Figure~\ref{fig:GeneralProblems}). The listed examples are problematic with respect to annotation and their prediction by our baselines -- due to the low IoU reported in the main paper. 

The images in Figure~\ref{fig:DiffCrack} display \textit{Cracks} that are bordered by strongly varying background. The most left image shows an abrupt change regarding the background at the vertical edge of an abutment wall, where one side represents a clean concrete surface and the other is heavily weathered (\textit{Weathering}). On the middle image, the \textit{Crack} traverses different colors of a \textit{Graffiti}. The most right sample displays vegetation which covers the underlying \textit{Crack}.
The two most left images in Figure~\ref{fig:BadPerformance} display defects that can be easily mixed up. 
\textit{Alligator Cracks} are areas of multiple \textit{Cracks} that are branched and arbitrary orientated, which makes their differentiation complicated. In the most left image, both of these classes are present. 
As mentioned in the main paper, differentiating between \textit{Weathering} and \textit{Wetspot} is problematic because they often overlap with each other (see middle image in Figure~\ref{fig:BadPerformance}). 
Another problematic class is \textit{Restformwork} which is under the lower five classes regarding the achieved IoU by our best model. In the most right image of Figure~\ref{fig:BadPerformance} \textit{Restformwork} is covered by concrete slurry. Furthermore, \textit{Restformwork} can be made of wood and the color of the polystyrene may vary. Thus, this class has many different visual appearances, which the model has to learn to summarize in one damage class.  
In Figure~\ref{fig:EasyMixedUp}, we display the three classes of dacl10k which are the most complicated to differ: \textit{Spalling}, \textit{Rockpocket} and \textit{Washouts/Concrete corrosion}. While they look very similar, their cause and assessment differ decisively (see Table~\ref{tab:ConcreteDefects}). 

Figure~\ref{fig:GeneralProblems} includes images representing class-independent difficulties (not exclusively for one class) and image quality problems across the dataset. 
Often, defects show no clear border, leading to inconsistent annotations and predictions. This ``smearing of damage borders'' can be observed for \textit{Efflorescence}, \textit{Weathering}, \textit{Wetspot} and \textit{Rust}. An example of the latter is shown in the most left image tile of Figure~\ref{fig:ClassProblems}. In addition, objects and defects can reach from the foreground to the background (see middle image in Figure~\ref{fig:ClassProblems}), and some defects may show weak differences to their surrounding area due to low contrast (see most right image in Figure~\ref{fig:ClassProblems}).     
Further problems concern the image quality, such as lens flares, partial overexposure due to the usage of flashlight or reflections (see Figure~\ref{fig:ImgQualProb}). 

\section{Baseline experiments}
\subsection{Additional training informations}
For each of the six combinations of encoder and semantic segmentation architecture, we examined a grid search for the learning rates $1e^{-4}$, $5e^{-4}$, $1e^{-3}$ and $5e^{-3}$ where the best value is chosen based on the validation loss. 

\subsection{Auxiliary multi-label results}
In Table~\ref{tab:test-results-aux} we display additional metrics from the auxiliary head of the best model on the test split. 

\begin{table}[ht]
\begin{center}
\footnotesize
\begin{tabular}{@{}lrrrr@{}}
\toprule
     Class    &   Precision  &  Recall & F1 Score &  \#images \\ \midrule
        Crack &      0.77    &  0.52   &   0.62   &    485 \\
       ACrack &      0.78    &  0.61   &   0.68   &     97 \\
Efflorescence &      0.82    &  0.60   &   0.69   &    460 \\
   Rockpocket &      0.68    &  0.55   &   0.60   &    529 \\
     WConccor &      0.42    &  0.24   &   0.31   &     33 \\
  Hollowareas &      0.84    &  0.77   &   0.81   &    312 \\
     Spalling &      0.80    &  0.76   &   0.78   &   1010 \\
 Restformwork &      0.69    &  0.51   &   0.59   &    251 \\ \midrule
      Wetspot &      0.59    &  0.42   &   0.49   &    298 \\
         Rust &      0.88    &  0.72   &   0.79   &    972 \\
     Graffiti &      0.87    &  0.67   &   0.76   &    235 \\
   Weathering &      0.75    &  0.68   &   0.72   &    916 \\ \midrule
ExposedRebars &      0.80    &  0.59   &   0.68   &    234 \\
      Bearing &      0.85    &  0.82   &   0.83   &    203 \\
       EJoint &      0.74    &  0.67   &   0.70   &     93 \\
     Drainage &      0.88    &  0.54   &   0.67   &    294 \\
   PEquipment &      0.88    &  0.81   &   0.84   &    396 \\
        JTape &      0.75    &  0.58   &   0.66   &    264 \\
\bottomrule
\end{tabular}
\caption{Test results for auxiliary head of the best model.} 
\label{tab:test-results-aux}
\end{center}
\end{table}

\section{Bridge inspection}
In the following, we describe the currently practiced bridge inspection process (analogue inspection), its limitations and the concept of the digitized inspection including the automated damage recognition enabled by multi-label semantic segmentation models. Thereby, we use the process of hands-on inspections in Germany as an example. It is important to note that inspection processes worldwide, and especially the defect documentation, are similar.

\subsection{Analogue inspection}
\begin{figure*}[ht]
    \begin{center}
    \includegraphics[width=.76\textwidth]{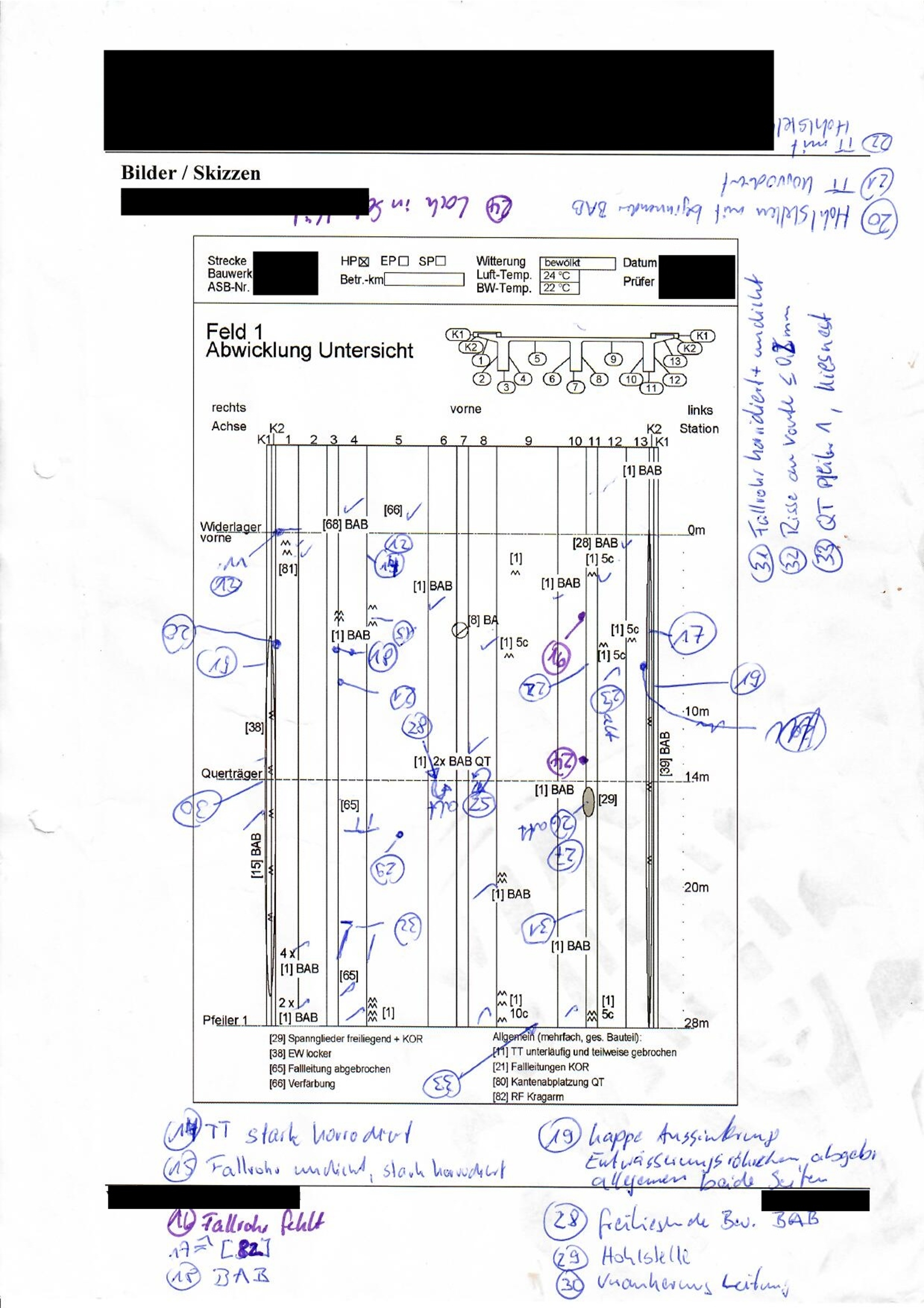}
    \caption{Damage sketch from a real bridge inspection with three T-beams. The sketch represents the bridge flipped-open along its longitudinal axis. Thus, the ``vertical lines'' represent the edges of the bridge cross-section. The defects are numbered in chronological order. Exemplary damage-information separated by comma (damage-number, damage-class, longitudinal, and cross-section dependent localization): 
    \#28, \textit{Spalling} (``BAB'') with \textit{Exposed Rebars} (``freiliegende Bewehrung''), at the cross girder (``Querträger''), on the left side of the web (6); 
    \#29, \textit{Hollowarea} (``Hohlstelle''), approximately 3m from the cross girder (``Querträger''), on the flange (5); 
    \#31, \textit{Drainage} (``Fallrohr'') with \textit{Rust} (``korrodiert'') which is leaking (``undicht''), approximately 7m from the cross girder (``Querträger''), on the bottom of the web (11).}
    \label{fig:DamageSketch}
    \end{center}
\end{figure*}

Currently practiced bridge inspections are carried out by a professionally trained civil engineer (bridge inspector). Usually, the inspector observes the complete surface of the given building while capturing each defect. During hands-on inspections, it is mandatory to detect visually recognizable defects and \textit{Hollowareas}, which can be detected by hammering the concrete surface. Thus, \textit{Hollowareas} are recognized based on the sound the hammering provokes\footnote{\textit{Hollowareas} and \textit{Cracks} are marked with chalk markings to make them easy to find in successive inspections.}. 
After the detection of a defect the inspector takes an image and notes, the damage class, size and location (damage-information) in a handwritten damage sketch. In Figure~\ref{fig:DamageSketch} such a sketch is displayed. In this case, the inspector numbered the defects chronologically and named the corresponding class and size (if necessary) of each damage, \eg,  \#32: Cracks (``Risse'') with a thickness of ``$\leq0.2 mm$'', approximately 7m from the cross girder (``Querträger''), between the left side of the web and flange of the most left T-beam (cross-section 4-5). After the inspection was completed, some inspectors additionally transfer the handwritten sketch to a CAD sketch. 

Finally, the assessment of the building is determined by calculating the condition grade, which is similar to grades in school, and preparing the inspection report. According to the damage-information noted in the damage sketch, the defect ID, which is defined by the German standard, can be assigned. For each defect and its ID, the country-specific standard recommends grades with respect to structural integrity, traffic safety and durability. Based on the expertise of the inspector, these recommendations are adapted. In the inspection report all defects are listed including their grades and recommendations regarding restoration, traffic load limitations or building a new construction. Furthermore, the defect sketch is attached to the report in order to enable a visual comparison with other -- especially consecutive -- inspections. This is important for tracking the defect development. 

\subsection{Limitations of the analogue inspection}
The process components of the analogue bridge inspection, such as classifying, measuring, locating and assessing the defects, is often inconsistent, error-prone, and lengthy. Oftentimes, this is due to the previously described cumbersome damage documentation.

According to Phares~\etal~\cite{Phares2004} the assessment results of bridge inspections vary greatly between inspectors. Thereby, Inspection reports from 49 bridge inspectors from 25 different state departments of transportation (DOTs) at seven structures, each in the United States, were evaluated. It was found that approximately 56\% of the condition ratings deviated significantly from the reference condition rating (ground truth). 
The main reasons for the strongly diverging ratings is the variability in inspection  documentation (e.g.,  field  inspection notes, photographs, etc.). Additionally, inspection notes concerning important structural defects or corresponding photographs were often omitted. 

We think, in addition, the following components of the analogue inspection contribute to the widely varying inspection results: 
\begin{itemize}
    \item The correct image must be found for the corresponding defect in the defect sketch after the inspection. This is problematic, especially on bridges with many defects that are close together. 
    \item Damage size is measured with a pocket rule, thus, it's imprecise. 
    \item Damage localization is often estimated or determined by measuring the distance to the next bridge pillar, which is inaccurate (or impossible) on bridges with a big span.
\end{itemize}

\subsection{Digitized inspection}
The basis for a digital building inspection (DI) is a Building Information Model (BIM), which is created during the building's planning phase or, in the case of existing bridges, before the inspection. At the structure the inspector records all defects making use of an UAV or smartphone. For non-hands-on inspections, UAVs can be used to record the defects because only visually recognizable defects have to be documented during this type of inspection. For hands-on (or close-up) inspections, smartphones or tablets are the most suitable devices to support the inspector, as they are handy and easy to use. In addition, close-up inspections require the detection of \textit{Hollowareas} which are detected by hammering the concrete surface, thus, making use of UAVs is not possible. Independent of the device, the automated damage detection is the central part of the inspection process, provided by a multi-label semantic segmentation model. The model enables the classification, measurement and localization on a pixel-level in order to assign an ID to each defect and to obtain the grading recommendations. These properties of each detected defect are stored as metadata for each DI. As a result, the BIM can be merged with the metadata of any performed inspection, allowing the generation of a digital twin. This allows for easily tracking the development of the bridge's condition over time. Consequently, instead of comparing damage sketches and inspection reports that are available in paper form, the evolution of the digital twin are assessed.  

The described DI has the goal to assist the inspector as much as possible in performing inspections and the consecutive evaluation within the framework of existing standards. The DI is simpler and more efficient than the analogue inspection, while the inspector remains the central decision-maker (human oversight). The acceleration of the inspection mainly results from the automated damage detection and the elimination of damage documentation on the defect sketch. 
For each detected damage, the DI-application provides a suggestion for the properties, or rather metadata. This recommendation can be corrected and saved by the inspector on site. This results in a direct quality management, which is particularly important for defects that require visual contextual knowledge.

\ctable[
star,
caption = {Concrete defects, their visual appearance, cause and example images with according annotation.},
label = tab:ConcreteDefects,
pos = htbp,
width=\textwidth,
doinside=\footnotesize,
captionskip=0.6ex,
]
{p{0.092\textwidth}m{0.25\textwidth}m{0.30\textwidth}m{0.269\textwidth}}
{}
{\FL
Damage & Visual appearance & Cause & Example \ML
Crack
    & 
    \begin{itemize}[leftmargin=*]
        \item Elongated and narrow zigzag line
        \item Clearly darker compared to the surrounding area or black
    \end{itemize}
    & 
    \begin{itemize}[leftmargin=*]
        \item Concrete’s tensile strength is exceeded
        \item Too high bending or shearing load 
        \item Settlement of the substructure 
    \end{itemize}
    &
    \includegraphics[height=4.7cm]{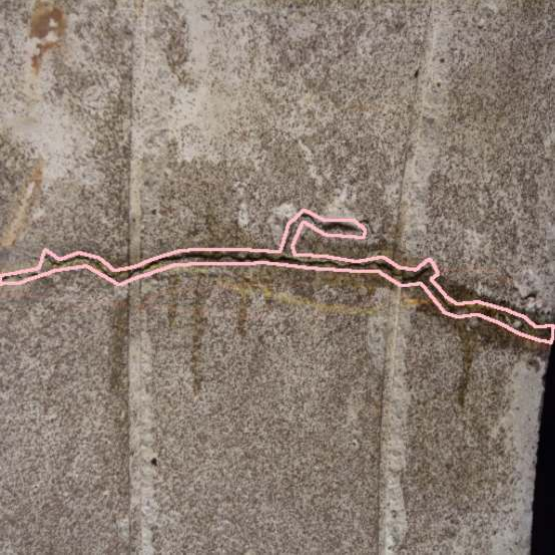} 
    \NN
    
Alligator Crack
(ACrack)
    & 
    \begin{itemize}[leftmargin=*]
        \item Many branched cracks
        \item Mostly arbitrarily orientated
        \item Usually with a small crack width (compared to \textit{Crack})
    \end{itemize}
    & 
    \begin{itemize}[leftmargin=*]
        \item Inadequate post-treatment or concrete recipe (shrinkage)
        \item Too high temperatures during hardening of the concrete
        \item Formation of expansive phases leading to a volume increase in the concrete as a result of chemical action
    \end{itemize}
    &
    \includegraphics[height=4.7cm]{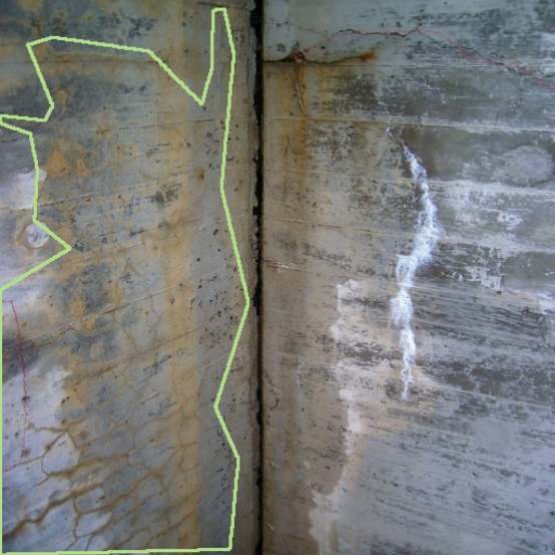} 
    \NN
       
Efflorescence
    & 
    \begin{itemize}[leftmargin=*]
        \item Mostly roundish areas of white to yellowish or reddish color
        \item Strong efflorescence can look similar to stalactites. 
        \item Often appears in weathered (\textit{Weathering}) or wet areas (\textit{WetSpot}) of the building and in combination with \textit{Crack} and/or \textit{Rust}
    \end{itemize}
    & 
    \begin{itemize}[leftmargin=*]
        \item Dissolving of salts (calcium, sodium, potassium) from the cement stone or aggregate by humidity changes or water ingress, \eg constantly running water through the building part or along its surface
        \item The salts consequently carbonate leading to the final visual appearance.
        \item Note: Water ingress can be caused by other defects imposing the draining of water, \eg, \textit{Restformwork} or damaged \textit{Drainage}. \textit{Efflorescences} are also called Calcium leaching.
    \end{itemize}
    &
    \includegraphics[height=4.7cm]{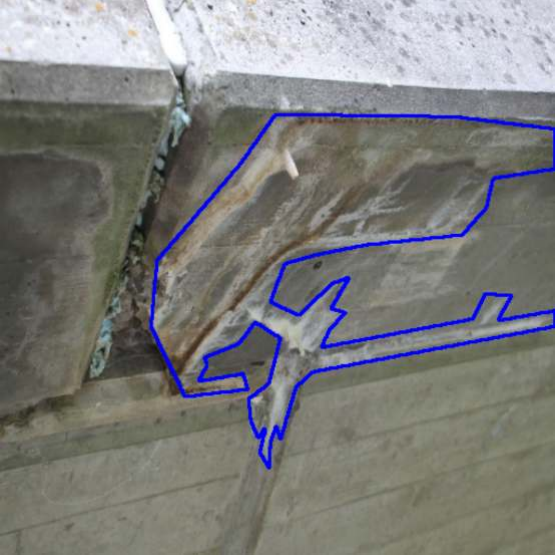} 
    \LL
}

\ctable[
star,
caption = {Concrete defects, their visual appearance, cause and example images with according annotation.},
pos = htbp,
width=\textwidth,
doinside=\footnotesize,
captionskip=0.6ex,
continued
]
{p{0.092\textwidth}m{0.25\textwidth}m{0.30\textwidth}m{0.269\textwidth}}
{
\tnote[1]{\textit{Cavity} was added in dacl10k\_v2 which was formerly included in \textit{Rockpocket}.}
}
{\FL
Damage & Visual appearance & Cause & Example \ML

Rockpocket
    & 
    \begin{itemize}[leftmargin=*]
        \item Visible coarse aggregate
        \item Often in tilts of the formwork and the bottom of building parts (opposite side from which the concrete is poured into the formwork) 
    \end{itemize}
    & 
    \begin{itemize}[leftmargin=*]  
        \item Inadequate rheological properties (viscosity, yield point) of the concrete 
        \item Appears due to bad compacting after having poured the concrete into the formwork.
        Therefore, insufficient deaeration of the concrete follows, causing areas where the cement paste didn't fill the volume between the coarse aggregate completely (\textit{Rockpocket}). 
        \item Note: Hence, the concrete cover of the reinforcement as well as the bond between the bars and the concrete is not provided or reduced. \textit{Rockpockets} are also called Honeycombing. 
    \end{itemize}
    &
 \includegraphics[height=4.7cm]{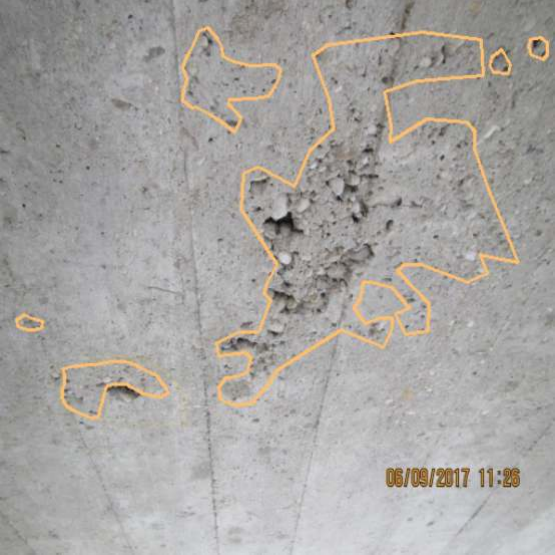} 
    \NN

Cavity\tmark[1]
    & 
    \begin{itemize}[leftmargin=*]
        \item Small air voids 
        \item Mostly on vertical surfaces 
    \end{itemize}
    & 
    \begin{itemize}[leftmargin=*]  
        \item Inadequate rheological properties (viscosity, yield point) of the concrete 
        \item Appears due to bad compacting after having poured the concrete into the formwork.
        Therefore, insufficient deaeration of the concrete follows causing small ''dots`` on the surface. 
        \item Note: At these spots the concrete cover is reduced. Cavities of usual size have no impact on the building assessment. 
    \end{itemize}
    &
 \includegraphics[height=4.7cm]{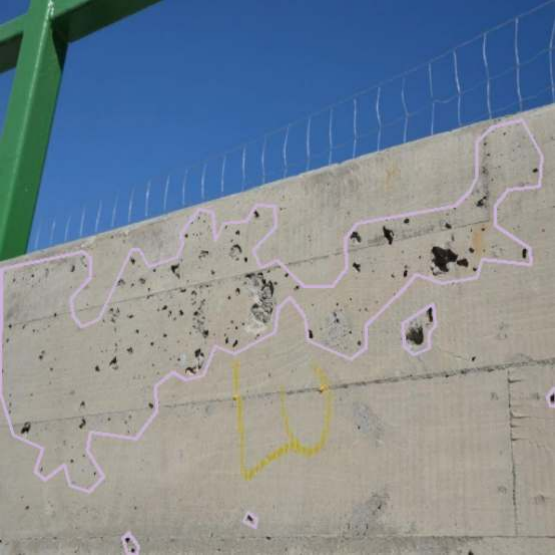} 
    \NN    

Concrete Corrosion
(ConcreteC)
    & 
    \begin{itemize}[leftmargin=*]
        \item Includes the visually similar defects: Washouts, Concrete corrosion and generally all kinds of planar corrosion/erosion/abrasion of concrete. 
        \item Note: We summarize all these ''planar corrosion defects`` in this class because they are visually hard to differ. According to inspection standards they have to subdivided which requires strong expertise in building defects.            
    \end{itemize}
    & 
    \begin{itemize}[leftmargin=*]
        \item Washouts appear on building parts that are constantly in contact with running water leading to the erosion of the concrete, \eg, abutment walls or bridge piers in rivers.   
        \item \textit{Concrete corrosion} can appear as a result of frost-thaw cycles, loss in succession to chemical attacks or abrasion (mechanical or action of acid and salt solutions). 
        \item Note: In dacl10k\_v1 \textit{Concrete Corrosion} (ConcreteC) was called ``Washouts/Concrete corrosion'' (WConccor). 
    \end{itemize}
    &
    \includegraphics[height=4.7cm]{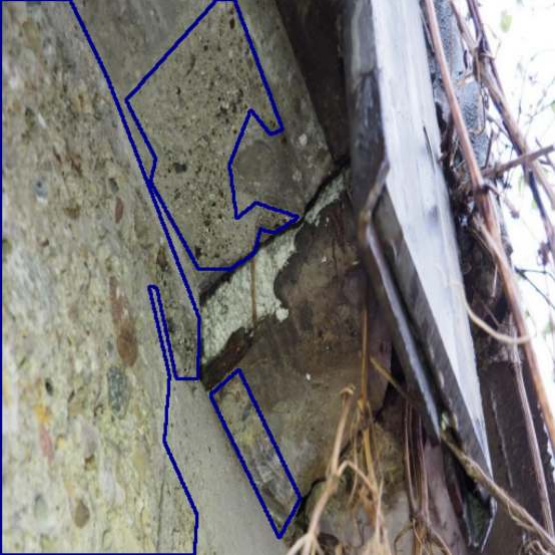} 
    \LL
}

\ctable[
star,
caption = {Concrete defects, their visual appearance, cause and example images with according annotation.},
pos = htbp,
width=\textwidth,
doinside=\footnotesize,
captionskip=0.6ex,
continued
]
{p{0.092\textwidth}m{0.25\textwidth}m{0.30\textwidth}m{0.269\textwidth}}
{}
{\FL
Damage & Visual appearance & Cause & Example \ML

Hollowarea
    & 
    \begin{itemize}[leftmargin=*]
        \item \textit{Hollowareas} are not visually recognizable but their markings made with crayons (mostly yellow, red or blue) during close-up/hands-on inspections.
        \item Note: The outer edge of the marking is considered as the boundary of the according area. We annotate every chalk marking that approximately forms a closed geometric figure. Single lines are not labeled as \textit{Hollowarea} as they are often used for the marking of \textit{Cracks}.   
    \end{itemize}
    & 
    \begin{itemize}[leftmargin=*]
        \item Corrosion of the subjacent reinforcement which leads to a volume increase surrounding the reinforcement bar and detaching of the concrete area that covers the bars. 
        \item \textit{Hollowareas} are usually the preliminary stage of \textit{Spallings} and \textit{Exposed Rebars}.
        \item Note: Recognizing \textit{Hollowareas} is very important as the falling concrete parts can cause severe damage. Therefore, hollow sounding areas are usually removed instantly. 
    \end{itemize}
    &
    \includegraphics[height=4.7cm]{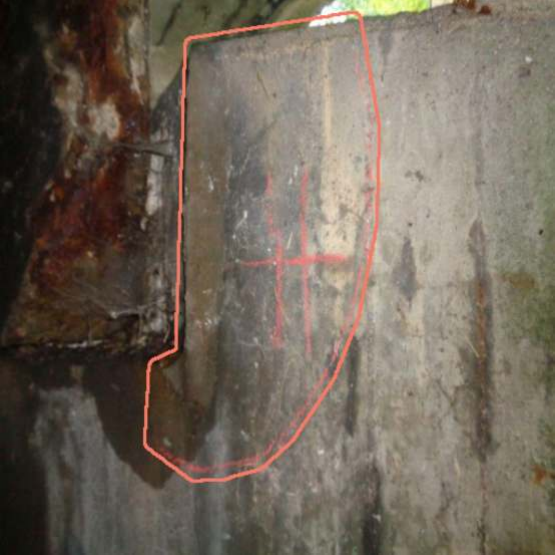} 
    \NN

Spalling
    & 
    \begin{itemize}[leftmargin=*]
        \item Spalled concrete area revealing the coarse aggregate
        \item Significantly rougher surface (texture) inside the \textit{Spalling} than in the surrounding surface
    \end{itemize}
    & 
    \begin{itemize}[leftmargin=*]
        \item Corrosion of the subjacent reinforcement leading to an increase in volume and consequent spalling of the concrete cover 
        \item Frost-thaw cycles of intruded water (deeper than \textit{Washouts/Concrete corrosion}) 
        \item Impact from vehicles, damaging during assembly of the building part or removal of the formwork, manufacturing faults    
    \end{itemize}
    &
    \includegraphics[height=4.7cm]{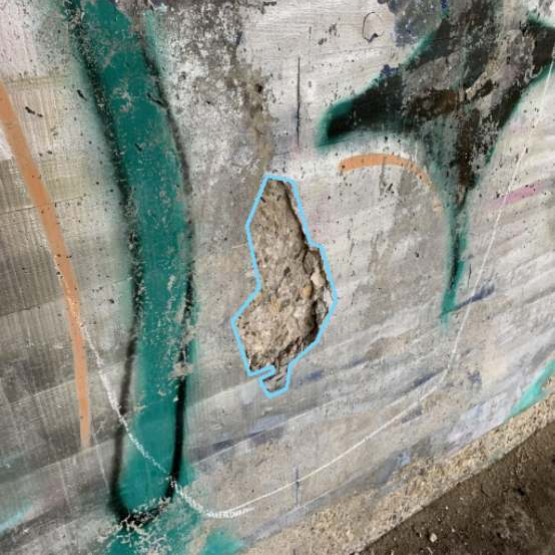} 
    \NN

Restformwork
    & 
    \begin{itemize}[leftmargin=*]
        \item Left pieces of formwork in joints or on the structure's surface
        \item Restformwork can be made of wood and polystyrene (PS).
        \item PS is often used as a placeholder in joints during concreting.
    \end{itemize}
    & 
    \begin{itemize}[leftmargin=*]
        \item After the concrete hardening has ended, PS is often forgotten to be removed (\eg, in the joint between the abutment wall and the superstructure).
        \item Note: If \textit{Restformwork} is not removed, water may be hindered to drain which can lead to other defects (\eg \textit{Spalling}, \textit{Exposed Rebars}, \textit{Rust}).
    \end{itemize}
    &
    \includegraphics[height=4.7cm]{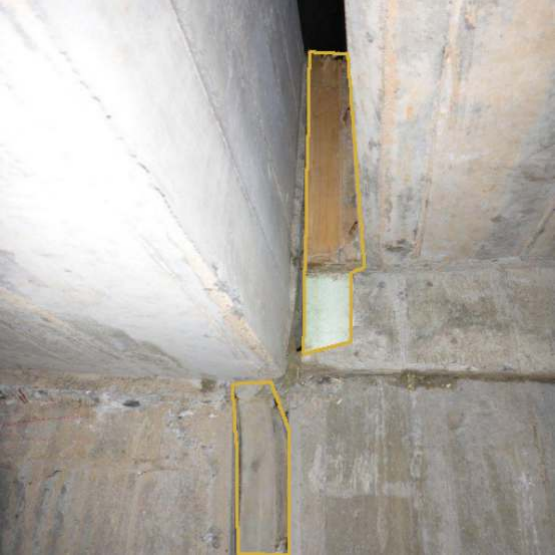} 
    \LL
}

\ctable[
star,
caption = {General defects, their visual appearance, cause and example images with according annotation.},
label = tab:GeneralDefects,
pos = htbp,
width=\textwidth,
doinside=\footnotesize,
captionskip=0.6ex,
]
{p{0.092\textwidth}m{0.25\textwidth}m{0.30\textwidth}m{0.269\textwidth}}
{}
{ \FL
Damage & Visual appearance & Cause & Example \ML
Wetspot
    & 
    \begin{itemize}[leftmargin=*]
        \item Wet/darker mirroring area
    \end{itemize}
    & 
    \begin{itemize}[leftmargin=*]
        \item Water is hindered to drain (through \textit{Restformwork}) or can't drain properly due to damaged \textit{Drainage}, leaky \textit{Expansion Joints}, \textit{Joint Tapes}, or \textit{Cracks} in the bridge deck.
        \item Note: There may be temporary \textit{Wetspots} due to recent rainfall which can be irrelevant for the bridge assessment. But, they may also indicate that the according area has to be observed in detail due to the greater exposition. In addition, the water can carry deicing salt (on road bridges). Usually, those areas are chosen to execute further investigations such as drill tests in order to determine the carbonation depth or chloride content.    
    \end{itemize}
    &
    \includegraphics[height=4.7cm]{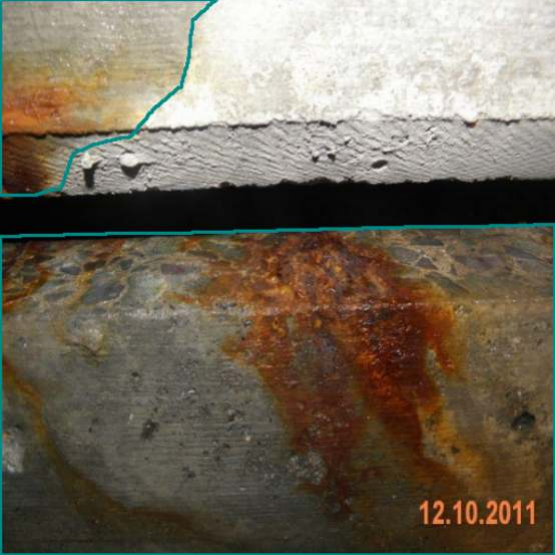} 
    \NN
    
Rust
    & 
    \begin{itemize}[leftmargin=*]
        \item Reddish to brownish area
        \item Often appears on concrete surfaces and metallic objects 
    \end{itemize}
    & 
    \begin{itemize}[leftmargin=*]
        \item Rust on the concrete surface originates from oxidation of the subjacent or neighboring reinforcement bars, or neighboring metallic building parts. The bars can corrode as a result of loss of the alkaline protective layer provided by "un-carbonated" concrete (pH value $>$ 9.5). If the pH value drops due to the further carbonation of the concrete (pH value $\leq$ 9.5), which is unavoidable over time, the reinforcement can oxidize.
        \item The carbonation is accelerated by \textit{Cracks} \textit{Rockpockets} and porous concrete because of faster intruding of water and carbon into the building part. In addition, the oxidation is intensified by deicing salts which is one of the most severe problems on road bridges. 
    \end{itemize}
    &
    \includegraphics[height=4.7cm]{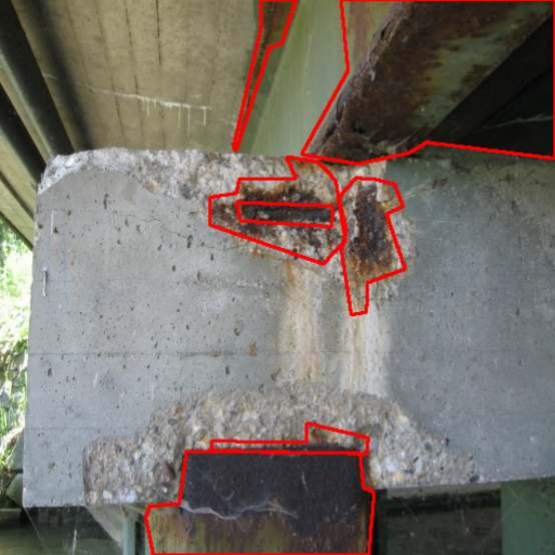} 
    \NN

Graffiti
    & 
    \begin{itemize}[leftmargin=*]
        \item All kinds of paintings on concrete and objects apart from defect markings
    \end{itemize}
    & 

    &
    \includegraphics[height=4.7cm]{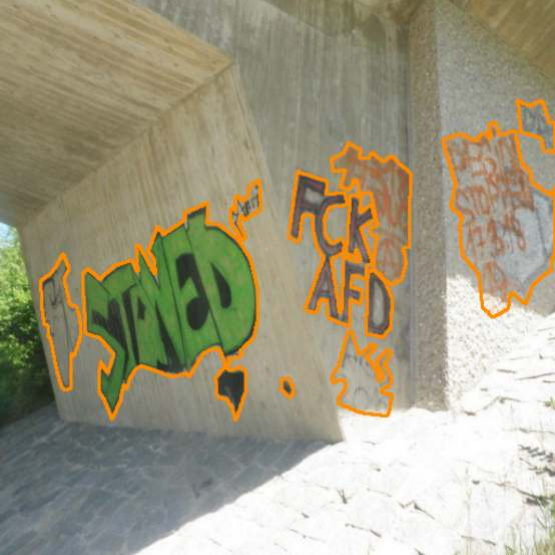} 
    \NN

Weathering
    & 
    \begin{itemize}[leftmargin=*]
        \item Summarizes all kinds of weathering on the structure (\eg smut, dirt, debris) and Vegetation (\eg plait, algae, moss, grass, plants).
        \item Weathering leads to a darker or greenish concrete surface compared to the rest of the surface. 
    \end{itemize}
    & 
    \begin{itemize}[leftmargin=*]
        \item Directly weathered areas
        \item Result of \textit{Wetspot}
        \item Note: \textit{Weathering} itself is not a severe defect. The main issue is that it can obscure other defects (\eg corroded reinforcement or cracks). \textit{Weathering} is also called Contamination.
    \end{itemize}
    &
    \includegraphics[height=4.7cm]{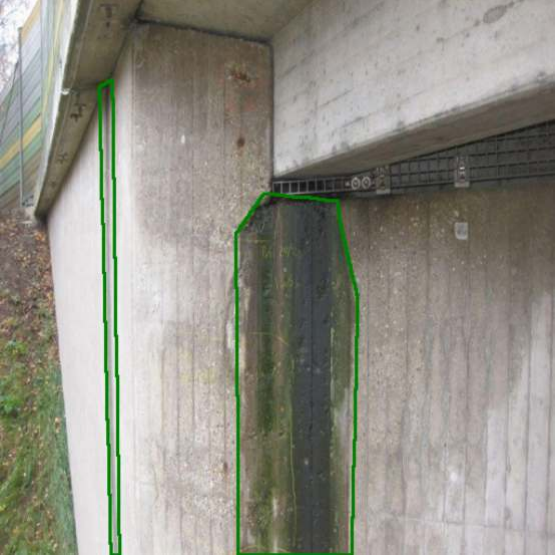} 
    \LL
}

\ctable[
star,
caption = {Objects, their visual appearance, functionality including their role for the bridge assessment and example images with according annotation.},
label = tab:Objects,
pos = htbp,
width=\textwidth,
doinside=\footnotesize,
captionskip=0.6ex,
]
{p{0.092\textwidth}m{0.25\textwidth}m{0.30\textwidth}m{0.269\textwidth}}
{}
{ \FL
Object & Visual appearance & Functionality & Example \ML
Exposed Rebars 
    & 
    \begin{itemize}[leftmargin=*]
        \item Exposed Reinforcement (non-prestressed and prestressed) and cladding tubes of tendons
        \item Often appears in combination with \textit{Spalling} or \textit{Rockpocket}, and \textit{Rust}
    \end{itemize}
    & 
    \begin{itemize}[leftmargin=*]
        \item In reinforced concrete structures, the reinforcement's task concerning the load transfer is to absorb the tensile forces.
        \item \textit{Exposed Rebars} occur due to insufficient concrete cover or corrosion and the consequent \textit{Spalling} of the concrete cover.
        \item The reduction of the reinforcement's cross section due to \textit{Rust} significantly influences the structural integrity of the according building.
    \end{itemize}
    &
    \includegraphics[height=4.7cm]{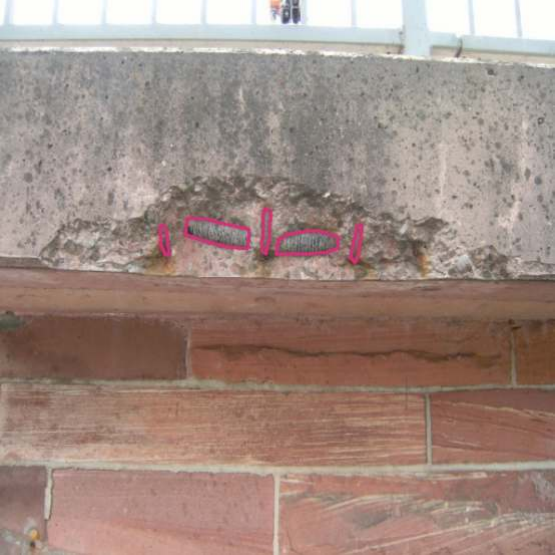} 
    \NN
    
Bearing
    & 
    \begin{itemize}[leftmargin=*]
        \item  All kinds of bearings, such as rocker-, elastomer- or spherical-bearings
    \end{itemize}
    & 
    \begin{itemize}[leftmargin=*]
        \item Bearings transfer the load from the superstructure to the substructure.
        \item Note: Bearings can show \textit{Rust} and \textit{Cracks} as well as deformation due to settlements in the abutments, overloading or creeping of the bridge. 
        The deformation or damaging of bearings can result in load redistribution which is not considered in the structural design of the bridge. 
    \end{itemize}
    &
    \includegraphics[height=4.7cm]{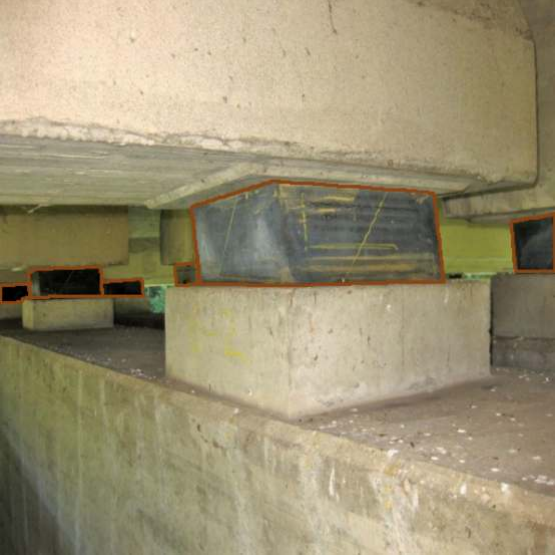} 
    \NN

Expansion Joint \linebreak
(EJoint)
    & 
    \begin{itemize}[leftmargin=*]
        \item Located at the beginning and end of the bridge
        \item Assembled cross to the longitudinal bridge axis
    \end{itemize}
    & 
    \begin{itemize}[leftmargin=*]
        \item \textit{Expansion Joints} compensate the thermal longitudinal expansion of the bridge deck and superstructure.
        \item Note: Mostly, \textit{Expansion Joints} are corroded (\textit{Rust}) and weathered (\textit{Weathering}) which hinders their ability to compensate the enlargements due to changes in temperature.
    \end{itemize}
    &
    \includegraphics[height=4.7cm]{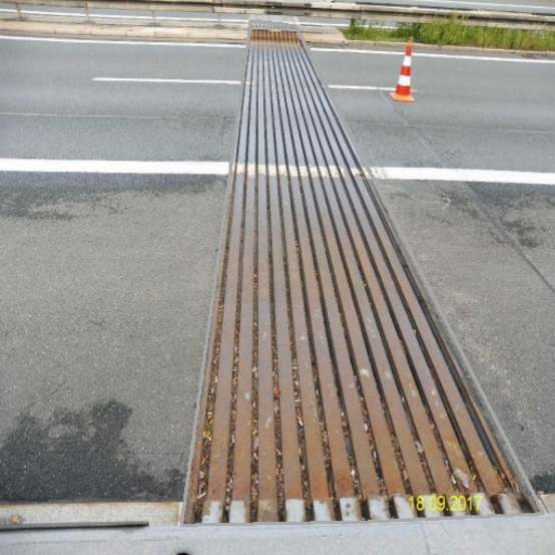} 
    \LL
}

\ctable[
star,
caption = {Object classes, their visual appearance, functionality including their role for the bridge assessment and example images with according annotation.},
pos = htbp,
width=\textwidth,
doinside=\footnotesize,
captionskip=0.6ex,
continued
]
{p{0.092\textwidth}m{0.25\textwidth}m{0.30\textwidth}m{0.269\textwidth}}
{}
{ \FL
Object & Visual appearance & Functionality & Example \ML
Drainage
    & 
    \begin{itemize}[leftmargin=*]
        \item  All kinds of pipes and outlets made of Polyvinylchlorid or metal mounted on the bridge.
    \end{itemize}
    & 
    \begin{itemize}[leftmargin=*]
        \item The \textit{Drainage} directs water (often contaminated with deicing salt) away from the bridge. 
        \item The draining of water is important for the durability of the bridge. If the water can't drain properly, especially when it's contaminated with deicing salts, the bridge's deterioration is accelerated leading to defects, such as \textit{Spalling}, \textit{Exposed Rebars} or \textit{Rust}
    \end{itemize}
    &
    \includegraphics[height=4.7cm]{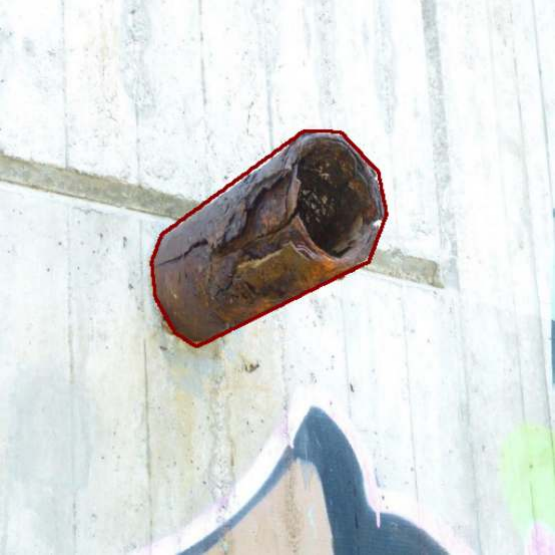} 
    \NN

Protective Equipment \linebreak
(PEquipment)
    & 
    \begin{itemize}[leftmargin=*]
        \item Railings, traffic safety features (\eg, steel rail, guide rails, impact attenuation device)
    \end{itemize}
    & 
    \begin{itemize}[leftmargin=*]
        \item Geometric adequacy and structural capacity of the \textit{Protective Equipment} is important with respect to the traffic safety. 
        \item The rail types and installation heights and minimum clearances must be checked.    
        \item Note: Mostly, they are corroded (\textit{Rust}) or deformed due to vehicle impact.
    \end{itemize}
    &
    \includegraphics[height=4.7cm]{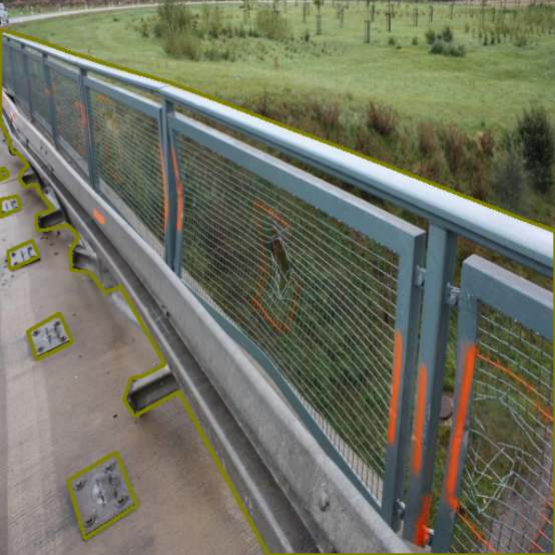} 
    \NN
    
Joint Tape \linebreak
(JTape)
    & 
    \begin{itemize}[leftmargin=*]
        \item  All joints that are filled with elastomer or silicon
        \item Note: Originally, \textit{Joint Tape} means an elastomer strap at the end and beginning of relatively small bridges. 
    \end{itemize}
    & 
    \begin{itemize}[leftmargin=*]
        \item \textit{Joint Tapes} compensate longitudinal enlargements due to changes in temperature (like \textit{Expansion Joint}).
        \item Note: A \textit{Joint Tape} is damaged when it's ruptured or twisted. \textit{Joint Tape} is also called Bridge Seal.
    \end{itemize}
    &
    \includegraphics[height=4.7cm]{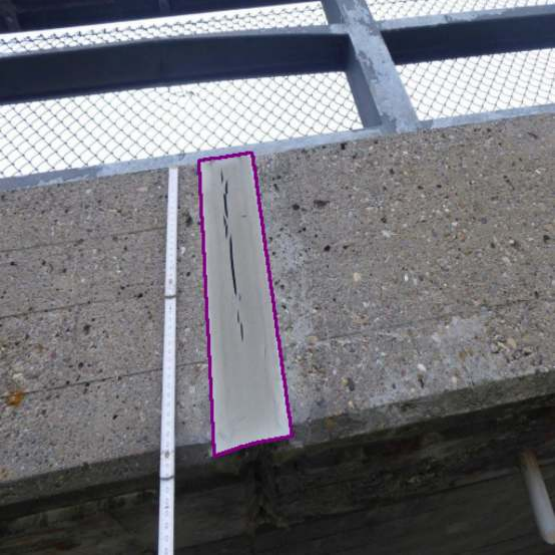} 
    \LL
}


\end{document}